\documentclass{article}

\PassOptionsToPackage{numbers, sort&compress}{natbib}
\PassOptionsToPackage{pagebackref,breaklinks,colorlinks}{hyperref}
\usepackage[main, final]{author-kit/neurips_2026}

\usepackage[utf8]{inputenc} %
\usepackage[T1]{fontenc}    %
\usepackage{hyperref}       %
\usepackage{url}            %
\usepackage{booktabs}       %
\usepackage{amsfonts}       %
\usepackage{nicefrac}       %
\usepackage{microtype}      %
\usepackage{xcolor}         %

\usepackage{amsmath,amsfonts,bm}

\newcommand{\captiona}{{(a)}\xspace}
\newcommand{\captionb}{{(b)}\xspace}
\newcommand{\captionc}{{(c)}\xspace}
\newcommand{\captiond}{{(d)}\xspace}

\newcommand{\refcaptiona}{{\textcolor{red}{a}}\xspace}
\newcommand{\refcaptionb}{{\textcolor{red}{b}}\xspace}
\newcommand{\refcaptionc}{{\textcolor{red}{c}}\xspace}

\def\eqref#1{equation~\ref{#1}}

\def\1{\bm{1}}

\def\rvd{{\mathbf{d}}}

\def\rvf{{\mathbf{f}}}
\def\rvg{{\mathbf{g}}}

\def\rvn{{\mathbf{n}}}
\def\rvo{{\mathbf{o}}}

\def\rvq{{\mathbf{q}}}
\def\rvr{{\mathbf{r}}}

\def\rvt{{\mathbf{t}}}

\def\vx{{\bm{x}}}

\def\evd{{d}}

\def\mA{{\bm{A}}}

\def\mI{{\bm{I}}}

\def\mP{{\bm{P}}}

\def\mR{{\bm{R}}}
\def\mS{{\bm{S}}}

\DeclareMathAlphabet{\mathsfit}{\encodingdefault}{\sfdefault}{m}{sl}
\SetMathAlphabet{\mathsfit}{bold}{\encodingdefault}{\sfdefault}{bx}{n}

\def\sM{{\mathbb{M}}}

\newcommand{\R}{\mathbb{R}}
\newcommand{\Manifold}{\mathbb{S}}

\newcommand{\normlone}{L^1}

\DeclareMathOperator*{\argmax}{\mathrm{arg\,max}}

\usepackage[numbers,sort&compress]{natbib}

\usepackage{caption}

\usepackage{titlecaps}  %

\usepackage{graphicx}
\usepackage{xspace}

\usepackage{lipsum}
\usepackage[mode=text, ]{siunitx}
\usepackage[dvipsnames]{xcolor}  %
\usepackage[cas-rgb*]{cncolours}

\usepackage{multirow}
\usepackage{makecell}  %
\usepackage{colortbl}

\usepackage{titlecaps}  %
\usepackage[capitalize]{cleveref}

\usepackage{pifont}  %
\usepackage{dsfont}  %

\usepackage{wrapfig}

\usepackage{tikz}

\usepackage[font=footnotesize,skip=3pt,subrefformat=parens]{subcaption}
\usepackage[normalem]{ulem}

\makeatletter
\renewcommand{\paragraph}{%
  \@startsection{paragraph}{4}{\z@}%
  {.1ex \@plus 0.1ex \@minus 0.0ex}%
  {-1em}%
  {\normalsize\bf}%
  }
  \makeatother

  \hypersetup{
    citecolor=Violet,
    urlcolor=Violet,
  }
  \newcommand{\CASE}[1]{{\MakeUppercase{#1}}}
  \newcommand{\enparen}[1]{\CASE{(}#1\CASE{)}}

  \colorlet{tableRankFirst}{朱红!35}
  \colorlet{tableRankSecond}{琥珀黄!35!white}
  \colorlet{tableRankThird}{芽绿!40}

  \newcommand{\TRankFirst}[1]{\cellcolor{tableRankFirst}\textbf{#1}}
  \newcommand{\TRankSecond}[1]{\cellcolor{tableRankSecond}\emph{#1}}
  \newcommand{\TRankThird}[1]{\cellcolor{tableRankThird}#1}

  \definecolor{tikzColorMethodGT}{HTML}{33a02c}
  \definecolor{tikzColorMethodOurs}{HTML}{1f78b4}
  \definecolor{tikzColorMethodOursParam}{HTML}{e05674}
  \definecolor{tikzColorMethodPlanaReLoc}{HTML}{ff7f00}
  \definecolor{tikzColorMethodGT}{HTML}{33a02c}
  \definecolor{tikzColorMethodGeoTr}{HTML}{cab2d6}
  \definecolor{tikzColorMethodMatchAny}{HTML}{a6cee3}
  \definecolor{tikzColorMethodNOPESAC}{HTML}{fb9a99}
  \definecolor{tikzColorMethodMast3r}{HTML}{ad9519}

  \definecolor{matchingOutliers}{HTML}{c01c28}
  \definecolor{poorPlaneRecovery}{HTML}{ffa348}

  \definecolor{surfaceMap}{HTML}{1c71d8}
  \definecolor{layoutMap}{HTML}{f36700}

  \definecolor{paramPath}{HTML}{c97a00}
  \definecolor{rayPath}{HTML}{ec4322}
  \definecolor{tokenPath}{HTML}{77767b}

  \crefname{section}{Sec.}{Secs.}
  \Crefname{section}{Section}{Sections}
  \Crefname{table}{Table}{Tables}
  \crefname{table}{Tab.}{Tabs.}
  \Crefname{equation}{Equation}{Equations}
  \crefname{equation}{Eq.}{Eqs.}
  \Crefname{appendix}{Appendix}{Appendices}
  \crefname{appendix}{Appendix}{Appendices}

  \newcommand{\qplanareloclegend}[1]{\raisebox{-.7ex}{\includegraphics[height=1.9\fontcharht\font`B]{assets/qualitative/planareloc-#1}}\xspace}
  \newcommand{\qfloclegend}[1]{\raisebox{-.7ex}{\includegraphics[height=1.9\fontcharht\font`B]{assets/qualitative/floc-#1}}\xspace}

\def\wo{\emph{w/o}\xspace}  %
\def\with{\emph{w/}\xspace}  %
\def\param{Param.%
}

\def\methodname{\textsc{Pggt}\xspace}

\def\planareloc{PlanaReLoc\xspace}
\def\spvloc{SPVLoc\xspace}

\def\mapname{dense surface map\xspace}
\def\mapnameplural{dense surface maps\xspace}
\def\layoutname{minimal layout model\xspace}
\def\layoutnameplural{minimal layout models\xspace}

\def\scannet{ScanNet\xspace}
\def\scannetpp{ScanNet++\xspace}
\def\matterport{Matterport3D\xspace}
\def\matterportsync{MP3D-Habitat\xspace}
\def\structured{Structured3D\xspace}
\def\twelvescenes{12Scenes\xspace}

\def\structuredofficial{S3D-\emph{official}\xspace}
\def\structuredresampled{S3D-\emph{resampled}\xspace}

\def\gtpose{the ground truth\xspace}

\def\renderandcompare{render-and-compare\xspace}

\makeatletter
\DeclareRobustCommand\onedot{\futurelet\@let@token\@onedot}
\def\@onedot{\ifx\@let@token.\else.\null\fi\xspace}

\def\eg{\emph{e.g}\onedot} 
\def\ie{\emph{i.e}\onedot}

\def\wrt{w.r.t\onedot}

\makeatother

\def\queryimage{\mI^{q}}
\def\querytokennum{l_q}
\def\scenemap{\sM}
\def\planarprimitive{\Pi}
\def\planarprimitivemap{\planarprimitive^{m}}
\def\mapplanenum{{l_m}}

\def\planeparameter{\bm{\pi}}
\def\planeparametermap{\planeparameter^{m}}
\def\planenormal{\rvn}
\def\planeoffset{\evd}
\def\planarprimitiveshape{\Omega}

\def\querypose{\rvg}
\def\queryposequaternion{\rvq}
\def\queryposetranslation{\rvt}
\def\canonicalposequaternion{\rvq_0}
\def\canonicalposetranslation{\rvt_0}

\def\RotError{\Delta \queryposequaternion}
\def\PosError{\Delta \queryposetranslation}

\def\tokendim{{c}}
\def\point{\vx}

\def\descriptor{\rvf}
\def\descriptorconfidence{\sigma}
\def\softassignment{\tilde{\mA}}
\def\assignment{\mA}
\def\similarity{\mS}
\def\matchthreshold{\tau}
\def\assignmentPos{\mathcal{A}^*}
\def\unassignablePatch{\mathcal{U}^q}
\def\unassignablePrimitive{\mathcal{U}^m}
\DeclareMathOperator*{\softmax}{\mathrm{Softmax}}

\def\parametermap{\mP}

\def\raymap{\mR}
\def\raymapuncertainty{\mR_u}

\def\raydirection{\rvd}
\def\rayorigin{\rvo}

\def\loss{{\mathcal{L}}}
\def\lossassignment{\loss_{\mathrm{assign}}}
\def\lossparameter{\loss_{\mathrm{param}}}
\def\lossray{\loss_{\mathrm{ray}}}
\def\losspose{\loss_{\mathrm{pose}}}
\def\lossseg{\loss_{\mathrm{seg}}}

\def\losscos{L_{\mathrm{cos}}}

\def\queryposetranslationTwoD{\queryposetranslation_{\mathrm{xy}}}
\def\queryposequaternionYaw{\queryposequaternion_{\mathrm{yaw}}}

\def\Tcheckmark{\ding{51}}
\def\TNoRankColor{gray!15}
\def\Thookrightarrow{{\footnotesize $\hookrightarrow$}\,}
\def\Tna{-}

\def\TMethods{Methods}
\def\TInferenceTime{Time\\\enparen{\unit{\second\per\text{iter}}} $\downarrow$}
\def\Tsim{~$\sim$~}
\def\Tdagger{\textsuperscript{$\dagger$}}

\def\TrenderAndCompare{Render~\&~\\Compare\ }
\def\TfeedForward{Feed-\\Forward}
\def\Tours{\textbf{Ours}}
\def\ToursSolver{\quad\Thookrightarrow \param}
\def\ToursToken{\quad\Thookrightarrow Token}
\def\ToursRay{\Tours~\enparen{Ray}}

\def\TParadigm{}

\def\TRotErr{$\RotError$ \enparen{\unit{\degree}} $\downarrow$}
\def\TPosErr{$\PosError$ \enparen{\unit{m}} $\downarrow$}
\def\TAccThres{Pose Recall \enparen{\unit{\percent}} $\uparrow$}

\def\TRotErrMean{\small Mean}
\def\TRotErrMed{\small Med.}
\def\TPosErrMean{\TRotErrMean}
\def\TPosErrMed{\TRotErrMed}

\def\AccFine{\enparen{\qty{0.2}{\meter},\,\qty{10}{\degree}}}
\def\AccMed{\enparen{\qty{0.5}{\meter},\,\qty{15}{\degree}}}
\def\AccCoarse{\enparen{\qty{1.0}{\meter},\,\qty{30}{\degree}}}

\def\TAccFine{\small \AccFine}
\def\TAccMed{\small \AccMed}
\def\TAccCoarse{\small \AccCoarse}

\def\TGeoTransformer{Geotransformer}

\def\Tfreereg{FreeReg}

\def\Tmastr{MASt3R}
\def\Tmatchanything{MatchAnything}

\def\Tnopesac{NOPE-SAC}
\def\Tplanerectr{PlaneRecTR}
\def\Tplanerectrpp{PlaneRecTR++}
\def\Tzeroplane{ZeroPlane}
\def\Tplanathreer{\textsc{Plana3R}\xspace}
\def\TplanarelocNoRefine{PlanaReLoc}
\def\Tplanareloc{\quad\Thookrightarrow +Refine}
\def\seqRANSAC{RANSAC\xspace}
\def\mogevtwo{MoGe-2\xspace}
\def\TMoGeRANSAC{\mogevtwo~\cite{wangMoGe2AccurateMonocular2025}{\scriptsize +\seqRANSAC}}

\def\TItoP{I2P}
\def\TMeshLoc{MeshLoc}
\def\TRM{RM}
\def\TOurs{}

\def\TMedianPoseErr{Med.
  Err.
  $\downarrow$}
\def\TMedianRotErr{$\RotError$ \enparen{\unit{\degree}}}
\def\TMedianPosErr{$\PosError$ \enparen{\unit{m}}}

\def\TMatchingPerformance{Primitive Matching~\enparen{\unit{\percent}}\,$\uparrow$}
\def\TMatchPrecision{\small \text{Prec.}}
\def\TMatchRecall{\small \text{Rec.}}
\def\TMatchFScore{\small \text{F$_\text{1}$}}
\def\TMatchAP{\small \text{AP}}

\def\TAblationMatchFScore{Matching \\ \TMatchFScore}

\def\TablationBase{\textbf{Full}~\enparen{Ray}}
\def\TablationNoAssign{\small\Thookrightarrow\wo $\lossassignment$}
\def\TablationNoParam{\small\Thookrightarrow\wo $\lossparameter$}
\def\TablationNoPose{\small\Thookrightarrow\wo $\losspose$}

\def\TablationNoPE{\small\Thookrightarrow\wo pos. enc.}
\def\TablationNoBoth{\small\Thookrightarrow\wo both}

\def\TFLocParadigm{Paradigm}

\def\TfLocRecall{Recall~\enparen{\unit{\percent}, 2D} $\uparrow$}
\def\TfLocPoseRecall{Pose \TfLocRecall}
\def\TfLocRecallFine{\qty{10}{\centi\meter}}
\def\TfLocRecallMed{\qty{50}{\centi\meter}}
\def\TfLocRecallCoarse{\qty{1}{\meter}}
\def\TfLocRTRecall{\qty{1}{\meter}~\&~\qty{30}{\degree}}

\def\TfLocPoseMedianError{$<$\qty{1}{\meter}
  Med.
  Error $\downarrow$~\enparen{2D}}
\def\TfLocPoseMedianErrorSixDoF{\TMedianPoseErr~\enparen{3D}}
\def\TMedianPosErrCM{$\PosError$ \enparen{\unit{cm}}}
\def\TfLocTransError{$\Delta \queryposetranslationTwoD$ \enparen{\unit{\centi\meter}}}
\def\TfLocRotError{$\Delta \queryposequaternionYaw$ \enparen{\unit{\degree}}}

\def\Tlaser{LASER}
\def\TfsanLoc{F$^\text{3}$Loc}
\def\TsanDP{3DP}
\def\TsemRayLoc{SemRayLoc}
\def\Trsk{RSK}
\def\TdicCoFLoc{DisCo}
\def\TUnLoc{UnLoc}
\def\TspvLoc{SPVLoc}

\def\TuseSemantics{Use \\Semantics}
\def\TreferenceModel{Map \\ Type}
\def\TfLocTwoD{2D}
\def\TfLocThreeD{3D}

\def\TFoVNinety{\qty{90}{\degree} FoV}
\def\TFoVOneTwenty{\qty{120}{\degree} FoV}
\def\Tquery{}

\def\TfLocAblationPoseMedianError{$<$\qty{1}{\meter}
  Med.
  Error $\downarrow$}
\def\TablationVanillaPGGT{Base~\enparen{Ray}}
\def\TablationPGGTSemantics{\small $+$ \with sem.
}
\def\TablationPGGTISOLATION{\small $+$ \with isolation}
\def\TablationPGGTFULL{\textbf{Full}~\enparen{Ray}}
\def\TAblationPoseRecall{Recall@\\ \qty{1}{\meter}~\&~\qty{30}{\degree} $\uparrow$}

\def\TDatasetMapType{Map\\Type}
\def\TDatasetName{Dataset}
\def\TDatasetDomain{Domain}

\def\TDatasetNumFrames{\#Frames}
\def\TDatasetCameraType{Camera\\Type}
\def\TDatasetNumScenes{\#Scenes\\~\enparen{regions}}
\def\TDatasetAvgFramesPerScene{Avg.\#Frames\\per Scene}

\def\DetailedMap{Surface}
\def\MinimalistMap{Layout}
\def\PerspectiveCamera{Perspective}
\def\PanoramicCamera{Panoramic}

\def\DatasetRealWorld{Real-world}
\def\DatasetSynthetic{Synthetic}

\def\TPlaneRecoverySegmentation{Segmentation}
\def\TSegmentationRI{RI $\uparrow$}
\def\TSegmentationVOI{VOI $\downarrow$}
\def\TSegmentationSC{SC $\uparrow$}

\def\TPlaneRecoveryDepth{Depth Estimation}
\def\TDepthRMSE{RMSE $\downarrow$}
\def\TDepthLog{Log\textsubscript{10} $\downarrow$}
\def\TDepthRel{Rel $\downarrow$}
\def\TDepthAccFine{$\sigma_1$ $\uparrow$}
\def\TDepthAccMed{$\sigma_2$ $\uparrow$}
\def\TDepthAccCoarse{$\sigma_3$ $\uparrow$}

\def\TSuppstructuredofficial{official}
\def\TSuppstructuredresampled{resampled-\TFoVNinety}
\def\TLayoutSegmentation{Segmentation $\uparrow$}
\def\TSegmentationmIoU{mIoU}
\def\TSegmentationmACC{mACC}

\title{\methodname-Loc: Primitive-Grounded Geometry Transformer for Feed-Forward Camera Localization}

\author{%
  \begin{minipage}[t]{\dimexpr\textwidth-2\tabcolsep\relax}
    \centering
    \textbf{Hanqiao Ye}\textsuperscript{\normalfont 1,2}\hspace{4em}
    \textbf{Yuzhou Liu}\textsuperscript{\normalfont 1,2}\hspace{4em}
    \textbf{Mengqi Rong}\textsuperscript{\normalfont 2}\hspace{4em}
    \textbf{Jian Liu}\textsuperscript{\normalfont 3}\hspace{4em}\\[.5pt]
    \textbf{Yangdong Liu}\textsuperscript{\normalfont  2}\footnotemark[1]\hspace{4em}
    \textbf{Shuhan Shen}\textsuperscript{\normalfont 1,2}\footnotemark[1]\\[2.5pt]
    \normalfont\small
    \textsuperscript{1}School of Artificial Intelligence, University of Chinese Academy of Sciences\\[.5pt]
    \textsuperscript{2}Institute of Automation, Chinese Academy of Sciences\hspace{2em}
    \textsuperscript{3}Beihang University\\[1.5pt]
    \texttt{\{yehanqiao2022, liuyuzhou2021, yangdong.liu\}@ia.ac.cn}\\[.5pt]
    \texttt{\{mengqi.rong,shshen\}@nlpr.ia.ac.cn}\hspace{2em}
    \texttt{liujian@lynmaxtech.com}
  \end{minipage}
}

\begin{document}

\maketitle

\footnotetext[1]{Corresponding authors.}
\vspace{-.2cm}
\begin{abstract}
  Camera localization methods are mostly multi-stage pipelines relying on dedicated and complex maps that are costly to construct, store, and maintain over time.
  Alternatively, well-structured maps that organize scenes with compact yet expressive primitives are often readily available, providing rich structural and semantic cues that humans can interpret.
  However, their potential for direct camera localization remains largely unexplored due to the significant modality gap between map primitives and visual observations.
  This motivates a new concept for localization, which estimates camera poses by \emph{grounding query observations to map primitives}.
  In this paper, we instantiate the concept in the context of planar primitives with \textbf{\methodname}, the \textbf{P}rimitive-\textbf{G}rounded \textbf{G}eometry \textbf{T}ransformer network.
  Given a single query image and a set of planar map primitives, the network is trained to predict multiple localization targets for feed-forward 6-DoF camera localization that generalizes to unseen environments.
  With this straightforward and scalable architecture, our method pushes the state-of-the-art localization performance on both dense surface maps and minimal layout models.
  The code and models are available at \url{https://github.com/3dv-casia/PGGT-Loc}.
\end{abstract}

\section{Introduction}

  Camera localization, the task of estimating the orientation and position of a query image relative to a pre-existing map, stands as a fundamental problem in computer vision and robotics, with wide applications in augmented reality~\enparen{AR}~\cite{zhaiSplatLoc3DGaussian2024}, robotic navigation~\cite{liActLocLearningLocalize2025}, and multi-agent collaboration~\cite{hanlonActiveVisualLocalization2024}.

  One popular family of approaches, known as structure-based localization,
  typically establishes 2D-3D correspondences between the query image and the scene structure, either by matching local features~\cite{camposecoHybridSceneCompression2019,sarlinCoarseFineRobust2019,humenbergerRobustImageRetrievalbased2022,liuEfficientGlobal2d3d2017,tairaInLocIndoorVisual2018,schonbergerSemanticVisualLocalization2018} or through direct scene coordinate regression~\cite{shottonSceneCoordinateRegression2013,brachmannDSACDifferentiableRANSAC2016,brachmannAcceleratedCoordinateEncoding2023,
    wangGLACEGlobalLocal2024,jiangRSCoReRevisitingScene2025}.
  With these explicitly established correspondences, the camera pose is then estimated by a robust geometric solver such as PnP-RANSAC~\cite{gaoCompleteSolutionClassification2003,fischlerRandomSampleConsensus1981,larssonPoseLibMinimalSolvers2020}.
  Note that these approaches require a dedicated map triangulated or trained from a large database of mapping images before a new image can be localized.
  While more general map representations have been explored for the task, \eg, meshes~\cite{panekMeshLocMeshbasedVisual2022} and NeRF/3DGS~\cite{yen-chenINeRFInvertingNeural2021,moreauCROSSFIRECameraRelocalization2023,chenRefinementAbsolutePose2024,huangSparseDenseCamera2025}, these methods still depend heavily on photorealistic scene appearance and often require an initial pose estimate for keypoint matching or iterative pose refinement.

  Well-structured maps are arrangements of high-level primitives and may further be enriched with relationships among them~\cite{armeni3DSceneGraph2019,zhengStructured3DLargePhotorealistic2020,avetisyanSceneCADPredictingObject2020,sarlinOrienterNetVisualLocalization2023,yang2024physcene,avetisyanSceneScriptReconstructingScenes2024,liuPolyRoomRoomawareTransformer2024,liuBuildingGPTAutoRegressiveBuilding2026}.
  Despite the significant modality gap between the map primitives and visual observations, which hinders the straightforward adaptation of the aforementioned localization paradigms, we argue that these map representations encode rich structural and even semantic information that remains underexploited for direct camera localization.

  \begin{figure}
    \centering
    \includegraphics[width=\linewidth]{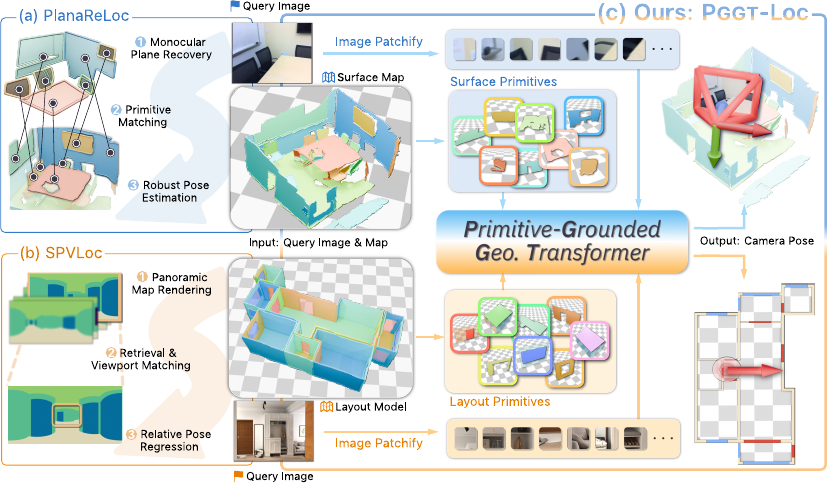}
    \caption{\textbf{Overview.}
      \captiona \planareloc~\cite{yePlanaReLocCameraRelocalization2026} localizes cameras against \emph{\mapnameplural} by explicitly matching 2D plane proposals with planar map primitives.
      \captionb \spvloc~\cite{gardSPVLocSemanticPanoramic2024} localizes cameras against \emph{\layoutnameplural} by retrieving the best-matching panoramic rendering.
      \captionc Our \methodname-Loc operates directly on query patches and map primitives, enabling feed-forward localization across both \emph{\mapnameplural} and \emph{\layoutnameplural} with a unified base architecture.
    }
    \label{fig:teaser}
    \vspace{-1.2em}
  \end{figure}

  Among the many map primitives, planes are particularly essential in representing physical surfaces, owing to their notable compactness and prevalence in structured environments~\cite{watsonAirPlanesAccuratePlane2024,xiePlanarReconRealtime3D2022,arcoreFundamentalConceptsEnvironmental2026}, which has motivated a recent work, named \planareloc~\cite{yePlanaReLocCameraRelocalization2026}, to explore their use for camera localization.
  As illustrated in \cref{fig:teaser}\refcaptiona, the method introduces a plane-centric feature-matching pipeline that first establishes cross-modal correspondences between planes in the query and the map, and then estimates the pose via 2D\textendash3D registration.
  While the method demonstrates the potential of planar primitives for streamlined camera localization without map texture, pose priors, or per-scene training, it still relies on an off-the-shelf plane recovery module to generate reliable 2D plane proposals and geometric priors for primitive matching and pose estimation, which poses a major bottleneck in practice.

  Meanwhile, the three-stage pipeline in \planareloc cannot be directly applied to \layoutnameplural~\cite{cruzZillowIndoorDataset2021,zhengStructured3DLargePhotorealistic2020}, whose primitives are restricted to architectural entities that do not span all physical surfaces.
  For localization on such minimalist maps, the \emph{\renderandcompare} paradigm dominates existing methods.
  As depicted in \cref{fig:teaser}\refcaptionb, the state-of-the-art method \spvloc~\cite{gardSPVLocSemanticPanoramic2024} first renders the layout map into a set of semantic panoramas by exhaustively sampling viewpoints across the scene, and then regresses the relative pose between the best-retrieved panorama and the query image.
  Although this paradigm neatly bridges the modality gap, reducing structure-based localization to matching 2D renderings inevitably discards structural information that could otherwise be exploited.

  In this work, motivated by the goal of fully leveraging high-level map primitives for more straightforward camera localization,
  we introduce \emph{primitive-grounded localization}, a new concept that casts the task as the grounding of query observations in map primitives.
  Driven by the concept and inspired by the DUSt3R~\cite{wangDUSt3RGeometric3D2024} architecture, we propose \textbf{\methodname}, a feed-forward network whose name stands for \textbf{P}rimitive-\textbf{G}rounded \textbf{G}eometry \textbf{T}ransformer.
  As shown in \cref{fig:teaser}\refcaptionc, \methodname takes a single query image and a set of map primitives as inputs, and predicts the camera pose \wrt the map in a single forward pass, eliminating the need to solve for the pose from explicit primitive correspondences or match the query against exhaustive 2D renderings of the map.
  We summarize our \textbf{contributions} as follows:
  \begin{itemize}
    \item We introduce \emph{primitive-grounded localization}, a new concept to localize cameras against well-structured maps through the grounding of query observations in map primitives.
    \item The concept is concretized in the context of planar primitives by \methodname, a cross-modal transformer network designed to ingest the query image and map primitives as direct inputs, and predict a set of localization-specific targets for feed-forward camera localization.
    \item \methodname provides a shared base architecture for localization tasks on both surface maps and layout models, surpassing prior works and demonstrating the potential of primitive-grounded localization as a promising paradigm for diverse primitive-based map representations.
  \end{itemize}

\section{Related Work}

  \paragraph{Localization with Higher-Level Maps.}
    Through computer vision, machines position themselves in the real world using diverse map representations, many of which are not encoded with visual observations but instead consist of high-level structural primitives and semantics interpretable by humans.
    In large outdoor scenarios, prior works have explored 2D semantic maps for 3-DoF localization, evolving from using only a few semantic classes~\cite{vojirEfficientLargescaleSemantic2020,Panphattarasap2018AutomatedMR,Guo2021CoarsetofineSL,Hong2019TextPlaceVP} to utilizing all available map elements~\cite{abonceYouAreHere2020,sarlinOrienterNetVisualLocalization2023,sarlinSNAPSelfsupervisedNeural2023,wuMapLocNetCoarsetofineFeature2024}.
    CAD/LoD models of urban buildings have also been used for 6-DoF localization~\cite{zhuLoDlocAerialVisual2024,panekVisualLocalizationUsing2023}.
    For indoor environments, 2D floorplan localization methods typically employ the \emph{\renderandcompare} paradigm, converting maps into diverse codebooks that facilitate intra-modal reasoning with query images~\cite{karkusParticleFilterNetworks2018,howard-jenkinsLalalocLatentLayout2021,minLASERLAtentSpacE2022,chenF$^3$LocFusionFiltering2024}.
    \planareloc~\cite{yePlanaReLocCameraRelocalization2026} and \spvloc~\cite{gardSPVLocSemanticPanoramic2024} exploit detailed planar surface maps and \layoutnameplural for 6-DoF camera localization, respectively, while adhering to conventional paradigms where primitive-encoded structural information remains insufficiently exploited.

  \paragraph{Multi-view Transformers.}
    DUSt3R~\cite{wangDUSt3RGeometric3D2024} comes with the two-view transformer architecture which comprises a shared ViT~\cite{dosovitskiyImageWorth16x162021} encoder and view-specific interconnected decoders, realizing feed-forward recovery of scene geometry and reshaping the landscape of 3D vision.
    Building upon this foundation, subsequent works have expanded the paradigm along multiple dimensions, such as multi-view inputs~\cite{linDepthAnything32026,wangVGGTVisualGeometry2025}, correspondence grounding~\cite{leroyGroundingImageMatching2024}, online streaming inference~\cite{muraiMASt3RSLAMRealTimeDense2025,wangContinuous3DPerception2025,zhuo2026streaming,chenTTT3R3DReconstruction2026}, more geometric inputs~\cite{jangPow3REmpoweringUnconstrained2025,keethaMapAnythingUniversalFeedForward2025}, dynamic scene modeling~\cite{zhangMonST3RSimpleApproach2025} and novel view synthesis~\cite{smart2024splatt3r,charatan23pixelsplat}.
    In particular, the paradigm also inspires several \emph{structureless} localization methods~\cite{dongReloc3rLargeScaleTraining2025,barroso-lagunaSceneWorthThousand2025}, which operate on groups of query and mapping images.
    In this work, we adapt DUSt3R's scalable architecture to \emph{cross-modal} inputs, aiming for feed-forward localization with high-level map primitives.

\section{Method}
  We propose \methodname, a cross-modal transformer network that instantiates the concept we propose: localizing by \emph{grounding query observations to map primitives}.
  \methodname is designed to ingest the query image and map primitives as direct inputs and predict multiple localization-specific targets, enabling feed-forward camera localization on well-structured maps that generalizes to unseen environments.

  \subsection{Problem Formulation}
    We unfold our method in the context of the \mapname $\scenemap$~\cite{monszpartRAPterRebuildingManmade2015,yuFindingGoodConfigurations2022,ye2025neuralplane,tanPlanarSplattingAccuratePlanar2025}, where a scene is represented as an inclusive collection of piecewise planar surfaces $\{\planarprimitivemap_i|i=1,\ldots,\mapplanenum\}$ with high geometric fidelity preserved.
    In the absence of additional information such as visual appearance, each map primitive $\planarprimitivemap_i\in \scenemap$ is fully determined by its plane parameter $\planeparametermap_i\in\Manifold^2\times\R$ and spatial extent $\planarprimitiveshape^m_i$~\enparen{represented by a boundary edge loop}.
    We use coefficients of the plane equation: $\planenormal\cdot\point=\planeoffset\,(\point\in\R^3)$ to define the plane parameter: $\planeparameter:=[\planenormal, \planeoffset]$, which is the concatenation of the plane normal $\planenormal\in\Manifold^2$ and the plane offset $\planeoffset\in\R$.

    Given a previously \emph{unseen} map $\scenemap$ and a query image $\queryimage\in\R^{3\times H\times W}$, the objective is to estimate the 6-DoF camera pose $\querypose\in\Manifold^3\times\R^3$ \wrt $\scenemap$ that corresponds to the query observation.
    We parametrize the camera pose as $\querypose:=[\queryposequaternion, \queryposetranslation]$, with $\queryposequaternion\in\Manifold^3$ and $\queryposetranslation\in\R^3$ denoting the rotation quaternion and translation vector, respectively.
    We assume that the query image is captured within the spatial extent of the map, with the observed region sufficiently mapped and exhibiting no major structural discrepancies with the map.
    In addition, our method does not necessarily require priors on camera intrinsics.

  \subsection{Model Architecture}
    \label{sec:model-architecture}

    \begin{figure}
      \centering
      \includegraphics{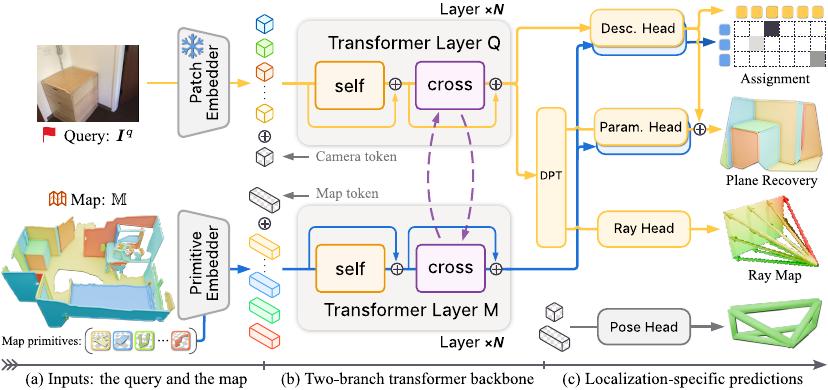}
      \caption{
        \textbf{Model Architecture}.
        \captiona Two embedders first tokenize the inputs into patch and primitive tokens, respectively.
        \captionb The two-branch transformer backbone refines the two sets of tokens, together with the prepended camera and map tokens, and enables interactions between them.
        \captionc Downstream prediction heads produce different localization targets that enable camera pose inference via three pathways, which we refer to as the \enparen{1} \param-based, \enparen{2}
        Ray-based, and \enparen{3} Token-based approaches.
      }
      \label{fig:architecture}
      \vspace{-1em}
    \end{figure}

    The \methodname architecture, illustrated in \cref{fig:architecture}, consists of two main components:
    \enparen{1} a two-branch transformer backbone consuming two modalities of inputs, namely the query and the map;
    \enparen{2} a set of output layers predicting multiple localization-specific targets.

    \paragraph{Cross-Modal Transformer Backbone.}
      In light of the tremendous potential of multi-view transformers in feed-forward 3D reconstruction, we inherit the architecture, adopting a backbone comprising two identical stacks of $N$ transformer decoder layers, with one stack processing query tokens and the other processing map tokens.
      Here, these two inputs are simply tokenized as follows:
      \begin{itemize}
        \item The query image $\queryimage$ is patchified into $\querytokennum$ tokens through a ViT encoder, which is pre-trained on large-scale image corpora and remains frozen throughout training.
        \item The map, represented as an unordered set of planar primitives $\{\planarprimitivemap_i|i=1,\ldots,\mapplanenum\}$, is encoded into $\mapplanenum$ tokens via a trainable projector, termed the primitive embedder.
              The primitive embedder tokenizes each map primitive through the integration of:
              \enparen{1} a vanilla 3D encoder that consumes points sampled on the primitive surface to capture its shape and spatial position;
              \enparen{2} a shallow MLP that projects the primitive parameter $\planeparametermap_i$ into a latent space, serving as the positional encoding.
      \end{itemize}

      Within each layer of the transformer backbone, the cross-attention block facilitates cross-modal interactions between the patch tokens and the primitive tokens.
      To predict dense outputs from the query image, a dedicated DPT layer is attached to the query branch, which converts patch tokens into pixel-aligned dense features.
      The DPT layer is instantiated with the Dual-DPT architecture proposed in DA3~\cite{linDepthAnything32026}: patch tokens collected from intermediate transformer layers are initially aggregated through shared reassembly modules, and then fused via separate fusion layers to produce task-specific feature maps.

    \paragraph{Prediction on Patch-Primitive Assignment.}
      An intuitive practice to propel the process of primitive grounding is to explicitly learn the unilateral assignment from image patches to map primitives.
      To this end, two distinct descriptor heads, one for each modality, project the backbone tokens into a shared feature space.
      Specifically, each token is projected by a 3-layer MLP into a descriptor $\descriptor\in\R^\tokendim$ and a confidence score $\descriptorconfidence\in[0,1]$.
      Each entry $\softassignment_{ij} (i\in[1, \querytokennum], j\in[1, \mapplanenum])$ of the soft assignment matrix, indicating the likelihood of the $i$-th image patch being matched/grounded to the $j$-th map primitive, is formulated as a combination of descriptor similarity $\similarity_{ij}$ and confidence scores:
      \begin{equation}
        \softassignment_{ij} = \descriptorconfidence^q_i\descriptorconfidence^m_j\cdot\softmax_{k\in[1, \mapplanenum]}\left(\similarity_{ik}\right)_j\text{, where }\similarity_{ik} = \descriptor^q_i\cdot\descriptor^m_k.
      \end{equation}
      Each image patch with a confidence score above $\matchthreshold$ is grounded to the primitive with the highest assignment score.
      Formally, the final assignment matrix is defined as $\hat{\assignment}_{ij}=1$ if {$j=\argmax_{k\in[1, \mapplanenum]}\softassignment_{ik}$} and $\descriptorconfidence^q_i>\matchthreshold$, indicating that the $i$-th patch is assigned to the $j$-th primitive, and $\hat{\assignment}_{ij}=0$ otherwise.

    \paragraph{Prediction on Primitive Parameters.}
      By predicting viewpoint-invariant 3D pointmaps, multi-view transformers iteratively recover scene structures and camera poses layer by layer~\cite{staryUnderstandingMultiViewTransformers2025}.
      Departing from their point-based representation, we further enhance our primitive grounding with 3D geometry learning by tasking our model to predict frame-invariant primitive parameters, enabling both local and mutual structure reasoning between the query and the map.
      In the query branch, one of the dense feature maps from the DPT layer is mapped into a per-pixel plane parameter map $\parametermap\in\R^{3\times H\times W}$ through a convolutional layer,
      while tokens from the map branch are decoded into plane parameters using a 3-layer MLP.
      Following \cite{yangRecovering3DPlanes2018,yuSingleimagePiecewisePlanar2019,liuInthewild3DPlane2025,shiPlaneRecTRUnifiedQuery2026}, the prediction of a plane parameter here is a 3D vector defined as $\planenormal/\planeoffset$.
      As map coordinate frames could be arbitrarily specified and may vary significantly across different input augmentations, we anchor all predictions of plane parameters in the query coordinate space, which provides a stable and consistent reference across different samples.
      Conceptually, this design encourages the query branch to focus on perceiving \emph{local} structure,
      while the map branch learns to align its primitives to the query view.

    \paragraph{Prediction on Camera Rays.}
      To directly target localization, a specialized prediction head is attached to the query branch to decode the dense features into a per-pixel ray map $\raymap\in\R^{6\times H\times W}$.
      Each camera ray $\rvr$ corresponds to the pixel through which it passes, and is defined as a 6D vector $[\raydirection; \rayorigin]$ where the ray direction $\raydirection\in\R^3$ and ray origin $\rayorigin\in\R^3$ are under the map coordinate frame.
      The ray map serves as an over-parameterized camera representation, implicitly encoding both the pose $\querypose$ and intrinsics.
      In the case of a classical perspective camera, the translation $\queryposetranslation$ is estimated by averaging the ray origins, whereas the intrinsics and rotation $\queryposequaternion$ are derived via decomposition of an optimal ray direction homography from an identity camera to the query camera, as detailed in \cite{linDepthAnything32026}.

    \paragraph{Prediction from Prepended Tokens.}
      To implement a more straightforward localization alternative while facilitating multi-task learning, we prepend a camera token to the query sequence and a map token to the map sequence.
      At the final backbone layer, the two tokens are decoded into corresponding poses through a shared MLP.
      In the map branch, the targeted pose is fixed to the identity pose with the rotation quaternion $\canonicalposequaternion=[0, 0, 0, 1]$ and translation $\canonicalposetranslation=[0, 0, 0]$, which represents the map's reference frame.
      Accordingly, the query pose in this reference frame can be directly decoded from the camera token.

    \paragraph{Summary.}
      Towards camera localization, \methodname predicts multiple localization targets in a single forward pass, including patch-primitive assignments, primitive parameters, camera rays, and poses.
      We then describe three pathways through which the pose of a query can be inferred:
      \enparen{1} solve a point-to-plane registration problem using the predicted patch-primitive assignment $\hat{\assignment}$, the predicted per-pixel parameters $\hat{\parametermap}$ and the given parameters of map primitives $\{\planeparametermap_i\}$, as in \cite{yePlanaReLocCameraRelocalization2026};
      \enparen{2} derive the pose from the predicted ray map $\hat{\raymap}$;
      \enparen{3} directly decode the pose from the camera token.
      We refer to the three alternatives above as \emph{\param}-based, \emph{Ray}-based and \emph{Token}-based approaches, respectively.

  \subsection{Training}
    \label{sec:methodology_training}

    \paragraph{Supervision.}
      \methodname is trained with a multi-task loss:
      \begin{equation}
        \loss = \lossassignment + \lossparameter + \lossray + \losspose,
        \label{equ:losses}
      \end{equation}
      which combines supervision signals on all the prediction targets.
      Specifically, $\lossassignment$ is the negative log-likelihood loss on the soft assignment $\softassignment$, with supervision from the ground-truth assignment $\assignment$;
      $\lossparameter$ combines the $\normlone$ loss and the cosine similarity loss applied to the predicted parameters;
      and both the ray map loss $\lossray$ and the pose decoding loss $\losspose$ are formulated using the $\normlone$ norm.

    \paragraph{Training Datasets.}
      Following the dataset processing workflow constructed by \cite{yePlanaReLocCameraRelocalization2026}, we scale our training dataset up to approximately 400K query-map pairs from three public datasets: \scannet~\cite{daiScanNetRichlyannotated3D2017}, \scannetpp~\cite{yeshwanthScanNetHighfidelityDataset2023} and \matterport~\cite{changMatterport3DLearningRGBD2017}, covering over four thousand indoor scenes/regions in total.
      Since \scannet is also used for evaluation, scenes for training and testing are strictly separated to ensure that all evaluation scenes are never seen during training.

    \paragraph{Implementation Details.}
      The patch embedder is instantiated with the DINOv2 ViT-Large~\cite{oquabDINOv2LearningRobust2023} backbone.
      In the primitive embedder, we employ PointNet~\cite{qiPointNetDeepLearning2017} as the 3D encoder, which takes \num{1024} points sampled from each primitive as input and is trained from scratch.
      The cross-modal transformer backbone is configured with $N=12$ layers for each modality and we feed patch tokens from the 2nd, 5th, 8th, and 11th layers into the DPT layer.
      The training images are resized and patchified into $\querytokennum$=1200 tokens.
      Maps are isotropically resized and centered into a unit box, and all predictions under the map coordinate frame should follow the same normalization.
      We use AdamW~\cite{loshchilovDecoupledWeightDecay2019} and a cosine LR schedule with a peak learning rate of \num{2e-4}.
      The model is trained for 125K iterations by default, with bfloat16 precision and a batch size of 128
      , which requires \num{\sim 2} days on 8$\times$H20 GPUs.

\section{Extension: Localization on 3D Layout}
  Having introduced \methodname in the \mapname, where query observations are grounded to map primitives that faithfully preserve the scene's physical surfaces, we note that the same paradigm naturally extends to minimalist 3D semantic models that represent only architectural layout.

  \paragraph{Task Setting.}
    The scene layout follows the annotation format of \structured~\cite{zhengStructured3DLargePhotorealistic2020}.
    As shown in \cref{fig:3d_floorplan}, it can be translated into a 2D floorplan or a 3D \layoutname that comprises a set of planar primitives, each
    corresponding to an architectural element.
    Compared with primitives in the \mapname, those in the \layoutname exhibit stronger regularity in terms of both their parameters and spatial extents, and each primitive is further associated with one of five distinct structural categories: wall, floor, door, window, and ceiling.
    Given such a \layoutname, we focus on localizing query images captured in a fully \emph{furnished} scene, where the increased modality gap poses a greater challenge to the detector-based feature matching conventions~\cite{yePlanaReLocCameraRelocalization2026}.

  \subsection{Model Adaptation}
    \label{sec:model-adaptions}
    \begin{wrapfigure}{r}{2.6in}
      \centering
      \includegraphics[width=\linewidth]{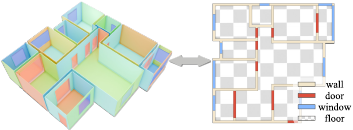}
      \caption{
        \textbf{Scene architectural layout.} \emph{Left}: 3D layout model with ceilings culled. \emph{Right}: corresponding 2D floorplan with semantic labels.
      }
      \label{fig:3d_floorplan}
      \vspace{-1em}
    \end{wrapfigure}
    Our base model performs reasonably well on the task.
    Here, we introduce two additional adaptations, highlighting its flexibility and potential in leveraging diverse map characteristics.

    \paragraph{Semantics Injection.}
      To integrate the provided primitive semantics, we initialize five learnable semantic embeddings within the primitive embedder.
      Each primitive token is enhanced with semantics by fusing its corresponding semantic embedding.
      We further supervise the query branch with an auxiliary semantic segmentation task after the DPT layer, fostering interactions between geometric and semantic representations.

    \paragraph{Isolation Injection.}
      Upon close inspection, as shown in \cref{fig:3d_floorplan}, the layout primitives are organized at the room level, \ie, no primitive is shared across rooms.
      Meanwhile, akin to prior works \cite{chenF$^3$LocFusionFiltering2024, gardSPVLocSemanticPanoramic2024}, we do not look into neighboring rooms through doors, treating all room connections as opaque.
      Consequently, query observations are expected to be grounded only to primitives within the same room.
      We enforce the isolation bias by applying attention masks to the self-attention blocks in the map backbone, ensuring that each primitive token attends only to other tokens within the same room.

  \subsection{Training}
    \label{sec:pggt-layout-training}
    The total loss in \cref{equ:losses} is augmented with the cross-entropy loss $\lossseg$ for semantic supervision.
    Following \cite{gardSPVLocSemanticPanoramic2024}, our model is trained from scratch on \num{3250} layout models from the training split of \structured~\cite{zhengStructured3DLargePhotorealistic2020}, with query images randomly sampled from \num{21835} photorealistic panoramic images.
    Both \emph{empty} and \emph{furnished} panoramas are used for training, and query images are sampled with FoVs ranging from \qty{45}{\degree} to \qty{135}{\degree}.
    Due to the shape regularity, the number of points sampled per primitive is reduced to \num{256}.
    Unless otherwise specified, the training recipe follows the settings in \cref{sec:methodology_training}.

\section{Experiments}
  \subsection{Results on \titlecap{\mapnameplural}}
    \label{sec:exp-planareloc}

    \paragraph{Evaluation Datasets.}
      Following the protocol and datasets proposed by \planareloc~\cite{yePlanaReLocCameraRelocalization2026}, we first conduct an evaluation of our method on plane-based surface maps using two public indoor datasets: \scannet~\cite{daiScanNetRichlyannotated3D2017} and \twelvescenes~\cite{valentinLearningNavigateEnergy2016}.
      The test set of \scannet consists of \num{7735} real-world images captured in \num{303} test scenes, while \num{1023} queries from \twelvescenes are used for out-of-domain evaluation.

    \paragraph{Evaluation Metrics.}
      To assess localization performance, we report results in terms of the mean/median localization errors: $\RotError$ and $\PosError$ for rotation and translation, respectively, along with pose recall, defined as the ratio of successful localizations under three thresholds.
      To assess how well the query observations are grounded to the map primitives, we transform the predicted patch-primitive assignment $\hat{\assignment}$ to 2D\textendash 3D plane correspondences by directly aggregating all patches assigned to the same primitive into a single 2D plane.
      Then, following \cite{yePlanaReLocCameraRelocalization2026}, we analyze the matching performance using Precision, Recall, F-score and Average Precision~\enparen{AP} at a mask IoU threshold of \num{0.3}.

      \begin{table}[htbp]
        \vspace{-.5em}
        \centering
        \caption{
          \textbf{Quantitative results} on \scannet.
          We report rotation and translation errors, along with pose recalls across three thresholds.
          We also report the average inference time per query, using a single NVIDIA RTX 3090 GPU.
          Top-3 results are highlighted as the {\setlength{\fboxsep}{1pt}\colorbox{tableRankFirst}{\textbf{first}}}, {\setlength{\fboxsep}{1pt}\colorbox{tableRankSecond}{\emph{second}}}, and {\setlength{\fboxsep}{1pt}\colorbox{tableRankThird}{third}}.
        }
        \vspace{.1em}
        \setlength{\tabcolsep}{6pt} %
        \renewcommand{\arraystretch}{1.2}
        \resizebox{\linewidth}{!}{%
          
\begin{tabular}{clcccccccc}
  \Xhline{3\arrayrulewidth}                                                                                                                                                                                                                                                                                                                                                                                                                                                                     \\[-1em]
                                                                     &                                                                  & \multicolumn{2}{c}{\TRotErr}  & \multicolumn{2}{c}{\TPosErr} & \multicolumn{3}{c}{\TAccThres} & \multirow{2}{*}{\rotatebox[origin=c]{0}{\makecell[t]{\TInferenceTime}}}                                                                                                                                                                               \\[0em]
  \cmidrule(r){3-4} \cmidrule(r){5-6} \cmidrule(r){7-9}                                                                                                                                                                                                                                                                                                                                                                                                                                         \\[-1.55em]
  \TParadigm                                                         & \TMethods                                                        & \TRotErrMean                  & \TRotErrMed                  & \TPosErrMean                   & \TPosErrMed                                                             & \TAccFine                     & \TAccMed                      & \TAccCoarse                   &                                                                             \\[0em]
  \Xhline{.3\arrayrulewidth}                                                                                                                                                                                                                                                                                                                                                                                                                                                                    \\[-1em]
  \multirow{5}{*}{\rotatebox[origin=t]{90}{\makecell[t]{\TMeshLoc}}} & \Tmastr~\cite{leroyGroundingImageMatching2024}                   & \tablenum{46.0}               & \tablenum{12.2}              & \tablenum{1.02}                & \tablenum{0.43}                                                         & \tablenum{35.4}               & \tablenum{49.5}               & \tablenum{57.6}               & \Tsim\tablenum[round-mode = places, round-precision=2]{0.683168}            \\[0em]
                                                                     & \Tmatchanything~\cite{heMatchAnythingUniversalCrossModality2025} & \tablenum{35.7}               & \tablenum{19.9}              & \tablenum{1.23}                & \tablenum{0.74}                                                         & \tablenum{20.0}               & \tablenum{35.7}               & \tablenum{52.1}               & \Tsim\tablenum[round-mode = places, round-precision=2]{0.925184}            \\[0em]
                                                                     & \Tnopesac~\cite{tanNOPESACNeuralOneplane2023}                    & \tablenum{28.7}               & \tablenum{15.9}              & \tablenum{0.90}                & \tablenum{0.77}                                                         & \tablenum{3.3}                & \tablenum{21.2}               & \tablenum{54.6}               & \Tsim\tablenum[round-mode = places, round-precision=2]{0.368222}            \\[0em]
                                                                     & \Tplanathreer~\cite{liuPLANA3RZeroshotMetric2025}                & \tablenum{26.8}               & \tablenum{12.9}              & \tablenum{0.92}                & \tablenum{0.52}                                                         & \tablenum{17.9}               & \tablenum{37.6}               & \tablenum{57.1}               & \Tsim\tablenum[round-mode = places, round-precision=2]{0.423516}            \\[0em]
                                                                     & \Tplanerectrpp~\cite{shiPlaneRecTRUnifiedQuery2026}              & \tablenum{20.7}               & \tablenum{9.9}               & \tablenum{0.66}                & \tablenum{0.54}                                                         & \tablenum{12.2}               & \tablenum{39.2}               & \tablenum{75.1}               & \Tsim\tablenum[round-mode = places, round-precision=2]{0.412271995}         \\[0em]
  \Xhline{.3\arrayrulewidth}                                                                                                                                                                                                                                                                                                                                                                                                                                                                    \\[-1em]
  \multirow{2}{*}{\rotatebox[origin=t]{90}{\makecell[t]{\TItoP}}}    & \TGeoTransformer~\cite{qinGeoTransformerFastRobust2023}          & \tablenum{45.2}               & \tablenum{26.5}              & \tablenum{1.42}                & \tablenum{1.06}                                                         & \tablenum{24.6}               & \tablenum{38.8}               & \tablenum{42.9}               & \Tsim\tablenum[round-mode = places, round-precision=2]{0.297699}            \\[0em]
                                                                     & \Tfreereg~\cite{wangFreeRegImagetoPointCloud2024}                & \tablenum{36.9}               & \tablenum{29.4}              & \tablenum{1.06}                & \tablenum{0.96}                                                         & \tablenum{0.8}                & \tablenum{6.4}                & \tablenum{33.1}               & \Tsim\tablenum[round-mode = places, round-precision=1]{14.187742}           \\[0em]
  \Xhline{.3\arrayrulewidth}                                                                                                                                                                                                                                                                                                                                                                                                                                                                    \\[-1em]
  \multirow{2}{*}{\rotatebox[origin=t]{90}{\makecell[t]{\TRM}}}      & \TplanarelocNoRefine~\cite{yePlanaReLocCameraRelocalization2026} & \TRankThird{\tablenum{17.3}}  & \TRankThird{\tablenum{3.9}}  & \TRankThird{\tablenum{0.65}}   & \TRankThird{\tablenum{0.27}}                                            & \TRankThird{\tablenum{37.1}}  & \TRankThird{\tablenum{69.8}}  & \TRankThird{\tablenum{79.8}}  & \Tsim{\bfseries\tablenum[round-mode = places, round-precision=2]{0.095019}} \\[0em]
                                                                     & \Tplanareloc~\cite{yePlanaReLocCameraRelocalization2026}         & \TRankSecond{\tablenum{17.2}} & \TRankSecond{\tablenum{3.8}} & \TRankSecond{\tablenum{0.60}}  & \TRankSecond{\tablenum{0.20}}                                           & \TRankSecond{\tablenum{48.5}} & \TRankSecond{\tablenum{73.1}} & \TRankSecond{\tablenum{81.8}} & \Tsim\tablenum[round-mode = places, round-precision=2]{0.544342}            \\[0em]
  \Xhline{.3\arrayrulewidth}                                                                                                                                                                                                                                                                                                                                                                                                                                                                    \\[-1em]
  \Xhline{.3\arrayrulewidth}                                                                                                                                                                                                                                                                                                                                                                                                                                                                    \\[-1.2em]
  \multirow{3}{*}{\rotatebox[origin=t]{90}{\makecell[t]{\TOurs}}}    & \ToursRay                                                        & \TRankFirst{\tablenum{7.5}}   & \TRankFirst{\tablenum{2.7}}  & \TRankFirst{\tablenum{0.25}}   & \TRankFirst{\tablenum{0.14}}                                            & \TRankFirst{\tablenum{69.0}}  & \TRankFirst{\tablenum{92.7}}  & \TRankFirst{\tablenum{95.4}}  & \Tsim\tablenum[round-mode = places, round-precision=2]{0.107205}            \\[0em]
  \rowcolor{\TNoRankColor}                                           & \ToursToken                                                      & \tablenum{7.8}                & \tablenum{2.9}               & \tablenum{0.26}                & \tablenum{0.14}                                                         & \tablenum{68.9}               & \tablenum{92.5}               & \tablenum{95.3}               & \Tna                                                                        \\[0em]
  \rowcolor{\TNoRankColor}                                           & \ToursSolver                                                     & \tablenum{8.6}                & \tablenum{2.5}               & \tablenum{0.40}                & \tablenum{0.16}                                                         & \tablenum{59.5}               & \tablenum{84.2}               & \tablenum{90.0}               & \Tna                                                                        \\[0em]
  \Xhline{3\arrayrulewidth}                                                                                                                                                                                                                                                                                                                                                                                                                                                                     \\[-1em]
\end{tabular}

        }
        \label{tab:exp_scannet}
        \vspace{-.8em}
      \end{table}

      \begin{table}[htbp]
        \centering
        \caption{
          \textbf{Quantitative results} on \twelvescenes.
          We present the cross-dataset performance in terms of both localization~\enparen{\ie, median errors and pose recall} and primitive matching.
        }
        \vspace{.1em}
        \setlength{\tabcolsep}{6pt} %
        \renewcommand{\arraystretch}{1.2}
        \resizebox{\linewidth}{!}{%
          
\begin{tabular}{lccccccccc}
  \Xhline{3\arrayrulewidth}                                                                                                                                                                                                                                                                                                                                                           \\[-1em]
                                                                   & \multicolumn{2}{c}{\TMedianPoseErr} & \multicolumn{3}{c}{\TAccThres} & \multicolumn{4}{c}{\TMatchingPerformance}                                                                                                                                                                                                 \\[0em]
  \cmidrule(r){2-3} \cmidrule(r){4-6} \cmidrule(r){7-10}                                                                                                                                                                                                                                                                                                                              \\[-1.55em]
  \TMethods                                                        & \TMedianRotErr                      & \TMedianPosErr                 & \TAccFine                                 & \TAccMed                      & \TAccCoarse                   & \TMatchPrecision              & \TMatchRecall                 & \TMatchFScore                 & \TMatchAP                     \\[0em]
  \Xhline{.3\arrayrulewidth}                                                                                                                                                                                                                                                                                                                                                          \\[-1em]
  \Tmastr~\cite{leroyGroundingImageMatching2024}                   & \tablenum{12.0}                     & \tablenum{0.30}                & \tablenum{45.2}                           & \tablenum{51.9}               & \tablenum{59.0}               & \TRankThird{\tablenum{59.8}}  & \tablenum{42.9}               & \tablenum{50.0}               & \TRankThird{\tablenum{81.6}}  \\[0em]
  \Tmatchanything~\cite{heMatchAnythingUniversalCrossModality2025} & \tablenum{7.9}                      & \TRankThird{\tablenum{0.20}}   & \TRankThird{\tablenum{46.9}}              & \tablenum{63.4}               & \tablenum{77.6}               & \tablenum{51.2}               & \TRankSecond{\tablenum{56.1}} & \TRankThird{\tablenum{53.5}}  & \tablenum{77.2}               \\[0em]
  \Tnopesac~\cite{tanNOPESACNeuralOneplane2023}                    & \tablenum{17.7}                     & \tablenum{0.54}                & \tablenum{2.9}                            & \tablenum{25.9}               & \tablenum{67.2}               & \tablenum{43.8}               & \tablenum{22.0}               & \tablenum{29.3}               & \tablenum{71.4}               \\[0em]
  \Tplanathreer~\cite{liuPLANA3RZeroshotMetric2025}                & \tablenum{15.9}                     & \tablenum{0.50}                & \tablenum{7.9}                            & \tablenum{30.1}               & \tablenum{69.3}               & \Tna                          & \Tna                          & \Tna                          & \Tna                          \\[0em]
  \Tplanerectrpp~\cite{shiPlaneRecTRUnifiedQuery2026}              & \tablenum{9.8}                      & \tablenum{0.39}                & \tablenum{11.1}                           & \tablenum{53.7}               & \TRankSecond{\tablenum{83.5}} & \tablenum{55.0}               & \tablenum{24.9}               & \tablenum{34.3}               & \tablenum{75.4}               \\[0em]
  \Xhline{.3\arrayrulewidth}                                                                                                                                                                                                                                                                                                                                                          \\[-1em]
  \TplanarelocNoRefine~\cite{yePlanaReLocCameraRelocalization2026} & \TRankThird{\tablenum{4.8}}         & \tablenum{0.28}                & \tablenum{34.9}                           & \TRankThird{\tablenum{66.7}}  & \tablenum{79.9}               & \TRankSecond{\tablenum{63.9}} & \TRankThird{\tablenum{54.2}}  & \TRankSecond{\tablenum{58.6}} & \TRankSecond{\tablenum{87.8}} \\[0em]
  \Tplanareloc~\cite{yePlanaReLocCameraRelocalization2026}         & \TRankSecond{\tablenum{4.7}}        & \TRankSecond{\tablenum{0.19}}  & \TRankSecond{\tablenum{50.6}}             & \TRankSecond{\tablenum{70.8}} & \TRankThird{\tablenum{80.6}}  & \Tna                          & \Tna                          & \Tna                          & \Tna                          \\[0em]
  \Xhline{.3\arrayrulewidth}                                                                                                                                                                                                                                                                                                                                                          \\[-1em]
  \Xhline{.3\arrayrulewidth}                                                                                                                                                                                                                                                                                                                                                          \\[-1.2em]
  \ToursRay                                                        & \TRankFirst{\tablenum{3.6}}         & \TRankFirst{\tablenum{0.15}}   & \TRankFirst{\tablenum{68.6}}              & \TRankFirst{\tablenum{90.1}}  & \TRankFirst{\tablenum{93.8}}  & \TRankFirst{\tablenum{77.7}}  & \TRankFirst{\tablenum{74.3}}  & \TRankFirst{\tablenum{76.0}}  & \TRankFirst{\tablenum{92.4}}  \\[0em]
  \Xhline{3\arrayrulewidth}                                                                                                                                                                                                                                                                                                                                                           \\[-1em]
\end{tabular}

        }
        \label{tab:exp_12scenes}
      \end{table}

\begin{figure}[htbp]
        \centering
        \begin{tikzpicture}
          \node[anchor=south west,inner sep=0] (image) at (0,0) {
            \includegraphics{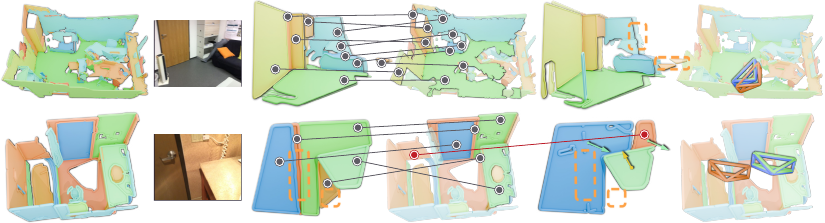}
          };

          {
          \def\subtitley{-0.08}
          \begin{scope}[x={(image.south east)},y={(image.north west)}]
            \node[fill=white, fill opacity=0.7, text opacity=1, inner sep=2pt] at (0.167, \subtitley) {\footnotesize \captiona Input: map \& query};
            \node[fill=white, fill opacity=0.7, text opacity=1, inner sep=2pt] at (0.474, \subtitley) {\footnotesize \captionb Plane recovery \& grounded primitives};
            \node[fill=white, fill opacity=0.7, text opacity=1, inner sep=2pt] at (0.73, \subtitley) {\footnotesize \captionc PlanaReLoc};
            \node[fill=white, fill opacity=0.7, text opacity=1, inner sep=2pt] at (0.91, \subtitley) {\footnotesize \captiond Localization};
          \end{scope}
          }
        \end{tikzpicture}
        \caption{
          \textbf{Qualitative comparisons with PlanaReLoc.}
          Our method robustly produces accurate pose estimation, with \captionb superior plane recovery and primitive correspondences as additional outputs.
          In contrast, \captionc \planareloc performs suboptimally and even fails due to poor plane recovery~\enparen{marked by the \textcolor{poorPlaneRecovery!
              90!black}{boxes}} and matching outliers~\enparen{bottom row, colored in \textcolor{matchingOutliers}{red}}.
          \captiond Comparison of localization results.
          Legend:
          \qplanareloclegend{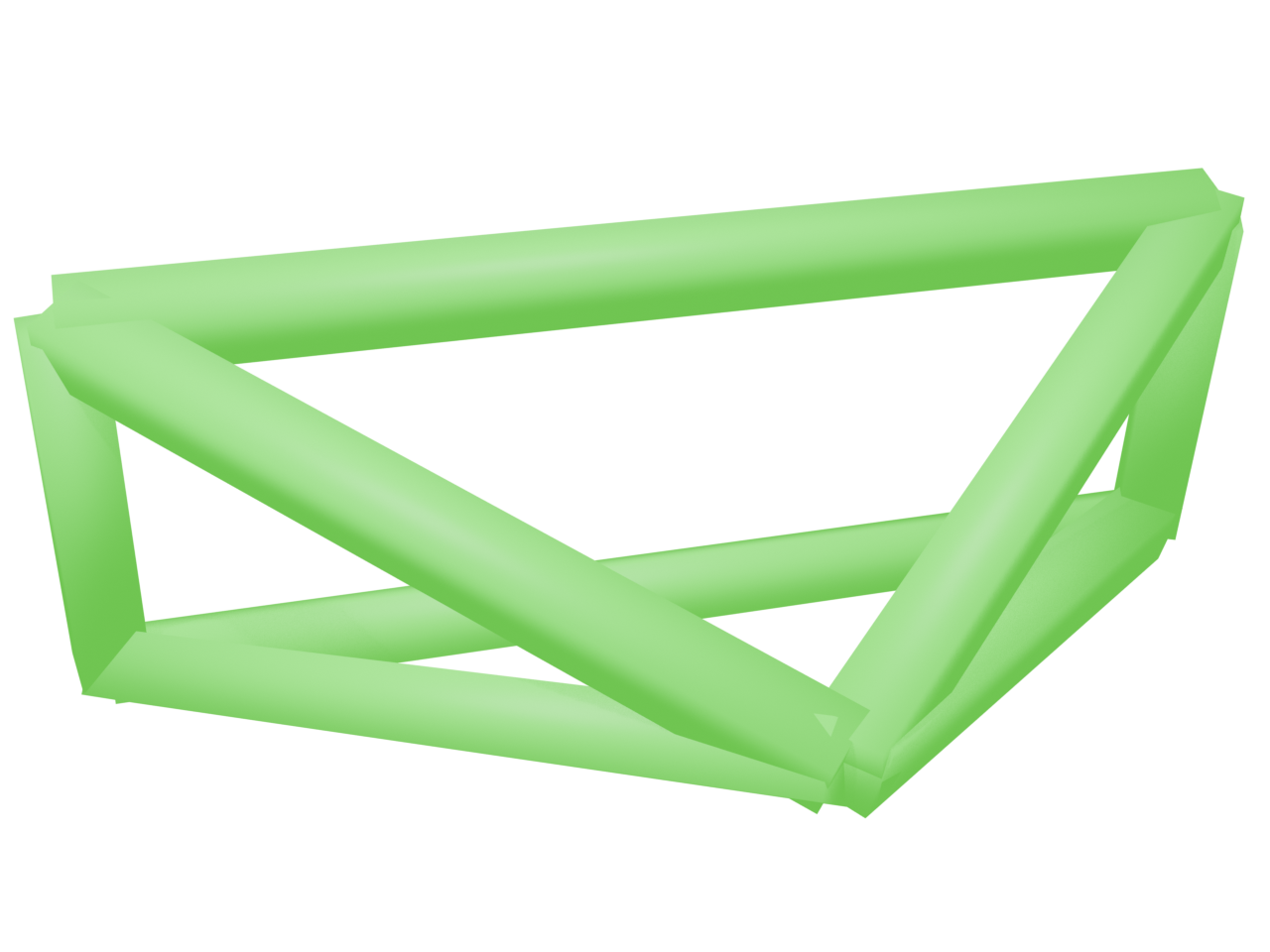}\textcolor{tikzColorMethodGT}{\gtpose},
          \qplanareloclegend{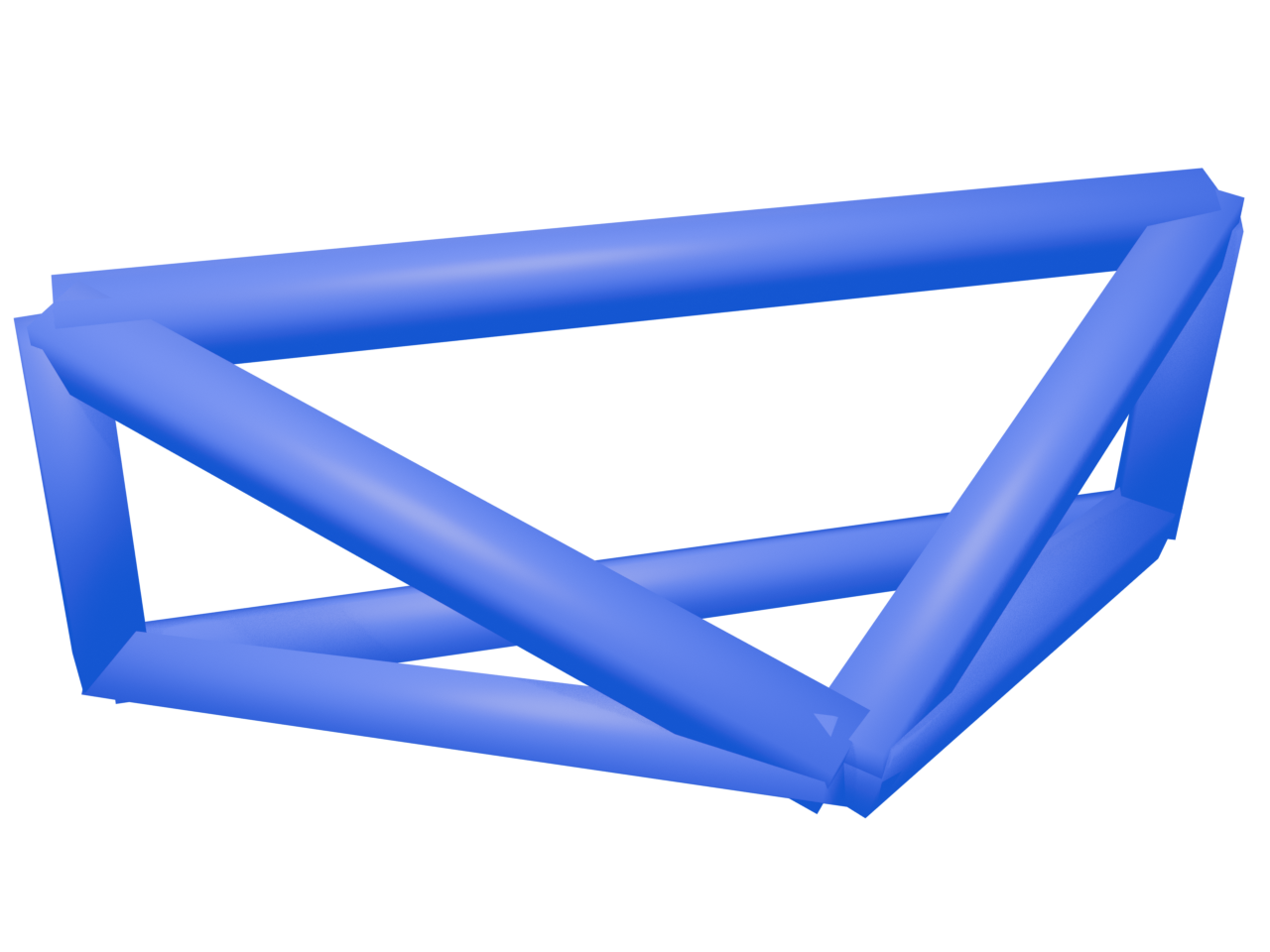}\textcolor{tikzColorMethodOurs}{\methodname\enparen{Ours}}, and
          \qplanareloclegend{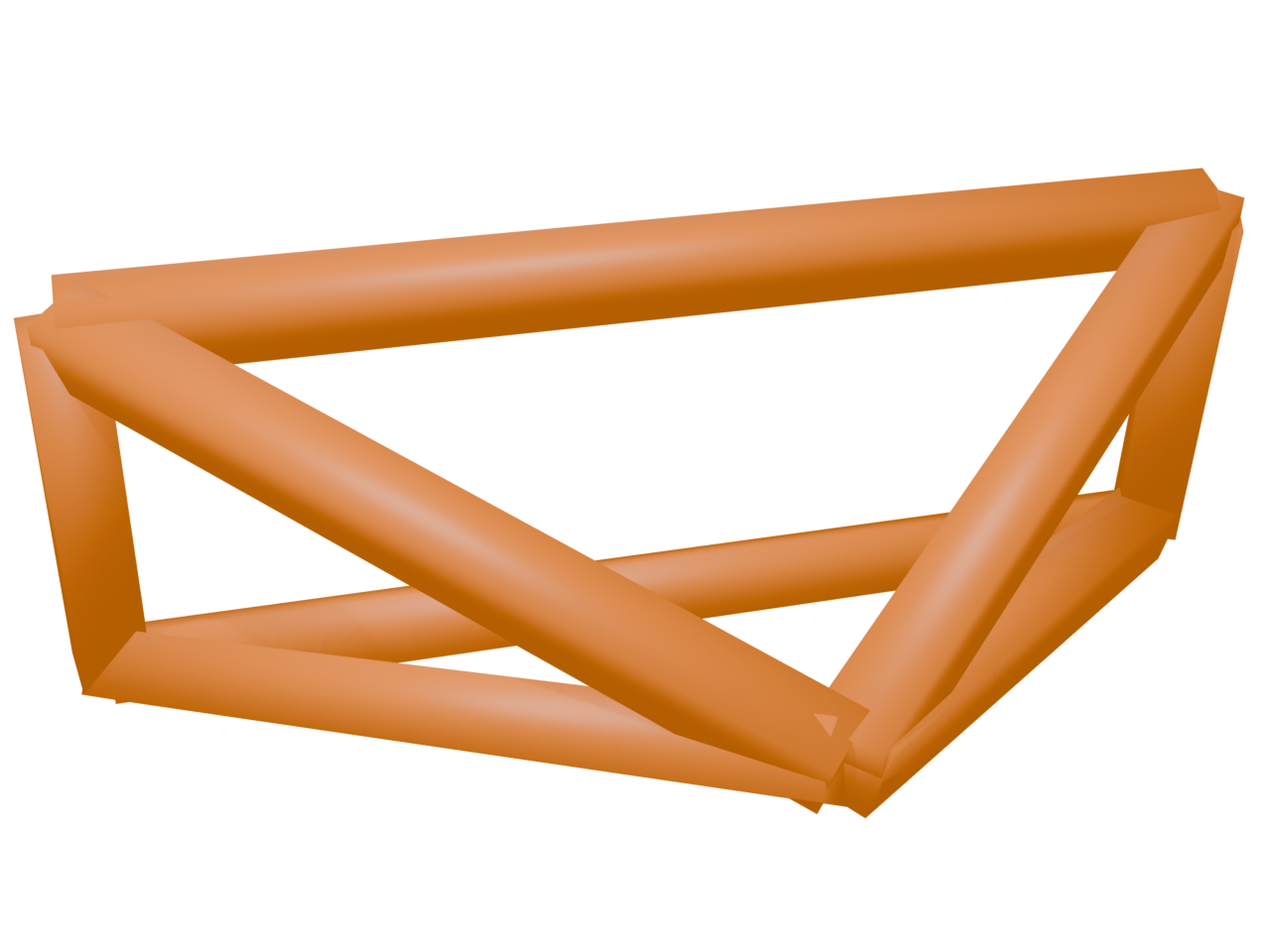}\textcolor{tikzColorMethodPlanaReLoc!
            80!black}{\planareloc}.
        }
        \label{fig:comparison_with_planareloc}
      \end{figure}

      Quantitative results on \scannet are presented in \cref{tab:exp_scannet}.
      We group our baselines into three categories:
      \enparen{1} \TMeshLoc~\cite{panekMeshLocMeshbasedVisual2022,panekVisualLocalizationUsing2023}, which localizes by estimating the relative pose between the query and the photorealistic rendering from the map, given a coarse initialization;
      \enparen{2} Image-to-Point cloud registration~\enparen{\TItoP}, which localizes by establishing 2D\textendash 3D point correspondences;
      \enparen{3} Region Matching~\enparen{\TRM}, which localizes by matching image regions to map primitives.
      The results of all baselines are taken from \cite{yePlanaReLocCameraRelocalization2026}.
      As the only feed-forward method capable of reconciling the query image and the structured map within a unified and scalable framework, our method achieves leading performance across all evaluation metrics.
      Particularly, poses derived from the ray map yield the best results, achieving an average relative improvement of \qty{28.6}{\percent} across the three pose recall metrics over \emph{\planareloc +Refine}.
      As reported in \cref{tab:exp_12scenes}, our method maintains strong performance in cross-dataset evaluation on \twelvescenes, while also achieving holistic improvements in primitive matching.
      \Cref{fig:comparison_with_planareloc} presents qualitative comparisons with \planareloc, where our method delivers more accurate and robust pose estimation, with superior plane recovery and primitive correspondences as by-products.

  \subsection{Results on \titlecap{\layoutnameplural}}
    \label{sec:exp-layout}

    \paragraph{Evaluation Datasets.}
      For the layout localization task, we evaluate our method on two datasets compiled from the test split of the \structured~\cite{zhengStructured3DLargePhotorealistic2020} dataset.
      The first, denoted as \structuredofficial, consists of \num{6405} high-resolution perspective query images from \num{250} scenes in the test split.
      The images are photo-realistically rendered with a 16:9 aspect ratio and a fixed horizontal FoV of \qty{80}{\degree}, including both empty and furnished scene configurations.
      The second, denoted as \structuredresampled, was introduced by \cite{gardSPVLocSemanticPanoramic2024}, in which the authors resampled query images from furnished RGB panoramas with controlled variations in FoV and observed semantics.
      We employ two subsets of \structuredresampled with FoVs of \qty{90}{\degree} and \qty{120}{\degree}, respectively, each containing \num{3426} square perspective images as queries.
      Note that before evaluation on \structuredofficial, we fine-tune our model using the officially provided perspective images from the training scenes, following \spvloc~\cite{gardSPVLocSemanticPanoramic2024}.

      \begin{table}[htbp]
        \centering
        \caption{
          \textbf{Quantitative results} on \structuredofficial.
          Different methods employ either 2D floorplans or 3D layout models as the reference map, with some further leveraging map semantics.
          We report median 2D localization errors over $<$\qty{1}{m} inliers, along with pose recall at different thresholds.
          Top-3 results are highlighted as the {\setlength{\fboxsep}{1pt}\colorbox{tableRankFirst}{\textbf{first}}}, {\setlength{\fboxsep}{1pt}\colorbox{tableRankSecond}{\emph{second}}}, and {\setlength{\fboxsep}{1pt}\colorbox{tableRankThird}{third}}.
          \Tdagger method not requiring known intrinsics~\enparen{\ie, FoV}.
        }
        \vspace{.1em}
        \setlength{\tabcolsep}{5pt} %
        \renewcommand{\arraystretch}{1.2}
        \resizebox{\linewidth}{!}{%
          
\begin{tabular}{clcccccccc}
  \Xhline{3\arrayrulewidth}                                                                                                                                                                                                                                                                                                                                                                                                                                                                                           \\[-1em]
                                                                             &                                                                 & \multirow{2}{*}{\rotatebox[origin=c]{0}{\makecell[t]{\TreferenceModel}}} & \multirow{2}{*}{\rotatebox[origin=c]{0}{\makecell[t]{\TuseSemantics}}} & \multicolumn{2}{c}{\TfLocPoseMedianError} & \multicolumn{4}{c}{\TfLocPoseRecall}                                                                                                                                 \\[0em]
  \cmidrule(r){5-6} \cmidrule(r){7-10}                                                                                                                                                                                                                                                                                                                                                                                                                                                                                \\[-1.55em]
  \TFLocParadigm                                                             & \TMethods                                                       &                                                                          &                                                                        & \TfLocTransError                          & \TfLocRotError                       & \TfLocRecallFine              & \TfLocRecallMed               & \TfLocRecallCoarse            & \TfLocRTRecall                \\[0em]
  \Xhline{.3\arrayrulewidth}                                                                                                                                                                                                                                                                                                                                                                                                                                                                                          \\[-1em]
  \multirow{8}{*}{\rotatebox[origin=t]{0}{\makecell[t]{\TrenderAndCompare}}} & \Tlaser~\cite{minLASERLAtentSpacE2022}                          & \TfLocTwoD                                                               &                                                                        & \Tna                                      & \Tna                                 & \tablenum{7.0}                & \tablenum{6.4}                & \tablenum{10.4}               & \tablenum{8.7}                \\[0em]
                                                                             & \TfsanLoc~\cite{chenF$^3$LocFusionFiltering2024}                & \TfLocTwoD                                                               &                                                                        & \Tna                                      & \Tna                                 & \tablenum{1.5}                & \tablenum{14.6}               & \tablenum{22.4}               & \tablenum{21.3}               \\[0em]
                                                                             & \TsanDP~\cite{chenPerspectiveHigherDimension2025}               & \TfLocTwoD                                                               &                                                                        & \Tna                                      & \Tna                                 & \tablenum{5.6}                & \tablenum{27.4}               & \tablenum{55.5}               & \tablenum{24.0}               \\[0em]
                                                                             & \TsemRayLoc~\cite{graderSuperchargingFloorplanLocalization2025} & \TfLocTwoD                                                               & \Tcheckmark                                                            & \Tna                                      & \Tna                                 & \tablenum{5.7}                & \tablenum{45.5}               & \tablenum{58.8}               & \tablenum{57.5}               \\[0em]
                                                                             & \Trsk~\cite{chenPerspectiveBroaderContext2025}                  & \TfLocTwoD                                                               &                                                                        & \Tna                                      & \Tna                                 & \tablenum{6.4}                & \tablenum{28.6}               & \tablenum{56.9}               & \tablenum{25.2}               \\[0em]
                                                                             & \TdicCoFLoc~\cite{meng2026discofloc}                            & \TfLocTwoD                                                               &                                                                        & \Tna                                      & \Tna                                 & \tablenum{10.0}               & \tablenum{59.0}               & \tablenum{67.0}               & \tablenum{66.0}               \\[0em]
                                                                             & \TUnLoc~\cite{wuestUnLocLeveragingDepth2026}                    & \TfLocTwoD                                                               &                                                                        & \Tna                                      & \Tna                                 & \tablenum{5.3}                & \tablenum{33.9}               & \tablenum{38.8}               & \tablenum{37.6}               \\[0em]
                                                                             & \TspvLoc~\cite{gardSPVLocSemanticPanoramic2024}                 & \TfLocThreeD                                                             & \Tcheckmark                                                            & \tablenum{12.9}                           & \tablenum{1.26}                      & \TRankThird{\tablenum{33.5}}  & \TRankThird{\tablenum{81.6}}  & \TRankThird{\tablenum{84.5}}  & \TRankThird{\tablenum{84.2}}  \\[0em]
  \Xhline{.3\arrayrulewidth}                                                                                                                                                                                                                                                                                                                                                                                                                                                                                          \\[-1em]
  \Xhline{.3\arrayrulewidth}                                                                                                                                                                                                                                                                                                                                                                                                                                                                                          \\[-1.2em]
  \multirow{3}{*}{\rotatebox[origin=t]{0}{\makecell[t]{\TfeedForward}}}      & \ToursRay\Tdagger                                               & \TfLocThreeD                                                             & \Tcheckmark                                                            & \tablenum{11.2}                           & \tablenum{0.42}                      & \TRankFirst{\tablenum{40.9}}  & \TRankFirst{\tablenum{91.3}}  & \TRankFirst{\tablenum{93.6}}  & \TRankFirst{\tablenum{93.5}}  \\[0em]
                                                                             & \ToursToken                                                     & \TfLocThreeD                                                             & \Tcheckmark                                                            & \tablenum{11.9}                           & \tablenum{0.46}                      & \TRankSecond{\tablenum{37.6}} & \TRankSecond{\tablenum{91.2}} & \TRankSecond{\tablenum{93.6}} & \TRankSecond{\tablenum{93.4}} \\[0em]
                                                                             & \ToursSolver                                                    & \TfLocThreeD                                                             & \Tcheckmark                                                            & \tablenum{14.4}                           & \tablenum{0.39}                      & \tablenum{27.5}               & \tablenum{60.5}               & \tablenum{76.7}               & \tablenum{76.0}               \\[0em]
  \Xhline{3\arrayrulewidth}                                                                                                                                                                                                                                                                                                                                                                                                                                                                                           \\[-1em]
\end{tabular}

        }
        \label{tab:exp_s3d_perspective169_door_closed}
        \vspace{-.5em}
      \end{table}

    \paragraph{Evaluation Metrics.}
      Since most prior studies on layout localization focus on estimating 3-DoF camera pose on 2D floorplans, we follow their standard evaluation protocol~\cite{minLASERLAtentSpacE2022,chenF$^3$LocFusionFiltering2024}.
      Specifically, we extract the 2D translation $\queryposetranslationTwoD\in\R^2$ and yaw angle $\queryposequaternionYaw\in\R$ from our 6-DoF pose predictions, from which we compute recall at multiple thresholds and report median errors over samples successfully localized within \qty{1}{m}~\enparen{\ie, $\Delta \queryposetranslationTwoD\le$\qty{1}{m}}.

      {
      \renewcommand{\TRankSecond}[1]{#1}
      \renewcommand{\TRankThird}[1]{#1}
      \begin{table}[t]
        \centering
        \caption{
          \textbf{Quantitative results} on \structuredresampled.
          For each FoV subset, we evaluate both 2D and 3D localization errors, along with 2D pose recall under different thresholds.
          The {\setlength{\fboxsep}{1pt}\colorbox{tableRankFirst}{\textbf{best}}} is highlighted.
        }
        \vspace{.1em}
        \setlength{\tabcolsep}{8pt} %
        \renewcommand{\arraystretch}{1.2}
        \resizebox{\linewidth}{!}{%
          
\begin{tabular}{clcccccccc}
  \Xhline{3\arrayrulewidth}                                                                                                                                                                                                                                                                                                                                                                                                      \\[-1em]
                                                                          &                                                 & \multicolumn{2}{c}{\TfLocPoseMedianErrorSixDoF} & \multicolumn{2}{c}{\TfLocPoseMedianError} & \multicolumn{4}{c}{\TfLocPoseRecall}                                                                                                                                                                 \\[0em]
  \cmidrule(r){3-4} \cmidrule(r){5-6} \cmidrule(r){7-10}                                                                                                                                                                                                                                                                                                                                                                         \\[-1.55em]
  \Tquery                                                                 & \TMethods                                       & \TMedianRotErr                                  & \TMedianPosErrCM                          & \TfLocRotError                       & \TfLocTransError              & \TfLocRecallFine              & \TfLocRecallMed               & \TfLocRecallCoarse            & \TfLocRTRecall                \\[0em]
  \Xhline{.3\arrayrulewidth}                                                                                                                                                                                                                                                                                                                                                                                                     \\[-1em]
  \multirow{5}{*}{\rotatebox[origin=t]{90}{\makecell[t]{\TFoVNinety}}}    & \Tlaser~\cite{minLASERLAtentSpacE2022}          & \Tna                                            & \Tna                                      & \tablenum{0.67}                      & \tablenum{14.6}               & \tablenum{23.5}               & \tablenum{72.8}               & \tablenum{76.8}               & \tablenum{76.2}               \\[0em]
                                                                          & \TspvLoc~\cite{gardSPVLocSemanticPanoramic2024} & \Tna                                            & \Tna                                      & \tablenum{0.83}                      & \tablenum{9.2}                & \tablenum{50.1}               & \TRankThird{\tablenum{89.7}}  & \TRankThird{\tablenum{92.1}}  & \TRankThird{\tablenum{91.9}}  \\[0em]
  \cmidrule{2-10}\noalign{\vskip -3pt}\cmidrule{2-10}
                                                                          & \ToursRay                                       & \TRankFirst{\tablenum{0.81}}                    & \TRankFirst{\tablenum{11.1}}              & \TRankSecond{\tablenum{0.38}}        & \TRankSecond{\tablenum{8.4}}  & \TRankFirst{\tablenum{56.0}}  & \TRankFirst{\tablenum{94.5}}  & \TRankFirst{\tablenum{95.6}}  & \TRankFirst{\tablenum{95.5}}  \\[0em]
                                                                          & \ToursToken                                     & \TRankThird{\tablenum{0.87}}                    & \TRankSecond{\tablenum{11.4}}             & \TRankThird{\tablenum{0.41}}         & \TRankThird{\tablenum{8.5}}   & \TRankThird{\tablenum{55.5}}  & \TRankSecond{\tablenum{94.4}} & \TRankSecond{\tablenum{95.6}} & \TRankSecond{\tablenum{95.4}} \\[0em]
                                                                          & \ToursSolver                                    & \TRankSecond{\tablenum{0.84}}                   & \TRankThird{\tablenum{11.5}}              & \TRankFirst{\tablenum{0.36}}         & \TRankFirst{\tablenum{7.6}}   & \TRankSecond{\tablenum{55.6}} & \tablenum{87.8}               & \tablenum{90.6}               & \tablenum{90.5}               \\[0em]
  \Xhline{.3\arrayrulewidth}                                                                                                                                                                                                                                                                                                                                                                                                     \\[-1em]
  \multirow{5}{*}{\rotatebox[origin=t]{90}{\makecell[t]{\TFoVOneTwenty}}} & \Tlaser~\cite{minLASERLAtentSpacE2022}          & \Tna                                            & \Tna                                      & \tablenum{0.76}                      & \tablenum{12.9}               & \tablenum{29.1}               & \tablenum{77.8}               & \tablenum{82.1}               & \tablenum{81.4}               \\[0em]
                                                                          & \TspvLoc~\cite{gardSPVLocSemanticPanoramic2024} & \Tna                                            & \Tna                                      & \tablenum{0.87}                      & \TRankFirst{\tablenum{11.7}}  & \TRankSecond{\tablenum{39.5}} & \TRankThird{\tablenum{88.0}}  & \TRankSecond{\tablenum{92.1}} & \TRankSecond{\tablenum{91.9}} \\[0em]
  \cmidrule{2-10}\noalign{\vskip -3pt}\cmidrule{2-10}
                                                                          & \ToursRay                                       & \TRankFirst{\tablenum{0.95}}                    & \TRankFirst{\tablenum{14.7}}              & \TRankFirst{\tablenum{0.49}}         & \TRankSecond{\tablenum{11.9}} & \TRankFirst{\tablenum{40.9}}  & \TRankFirst{\tablenum{91.7}}  & \TRankFirst{\tablenum{96.4}}  & \TRankFirst{\tablenum{96.3}}  \\[0em]
                                                                          & \ToursToken                                     & \TRankSecond{\tablenum{1.00}}                   & \TRankSecond{\tablenum{15.5}}             & \TRankThird{\tablenum{0.54}}         & \tablenum{12.8}               & \TRankThird{\tablenum{37.7}}  & \TRankSecond{\tablenum{91.6}} & \TRankFirst{\tablenum{96.4}}  & \TRankFirst{\tablenum{96.3}}  \\[0em]
                                                                          & \ToursSolver                                    & \TRankThird{\tablenum{1.11}}                    & \TRankThird{\tablenum{16.8}}              & \TRankSecond{\tablenum{0.54}}        & \TRankThird{\tablenum{12.0}}  & \tablenum{36.7}               & \tablenum{83.2}               & \TRankThird{\tablenum{87.0}}  & \TRankThird{\tablenum{86.2}}  \\[0em]
  \Xhline{3\arrayrulewidth}                                                                                                                                                                                                                                                                                                                                                                                                      \\[-1em]
\end{tabular}

        }
        \label{tab:exp_s3d_perspective11}
        \vspace{-.2em}
      \end{table}
      }

\begin{figure}[htbp]
        \centering
        \begin{tikzpicture}
          \node[anchor=south west,inner sep=0] (image) at (0,0) {
            \includegraphics{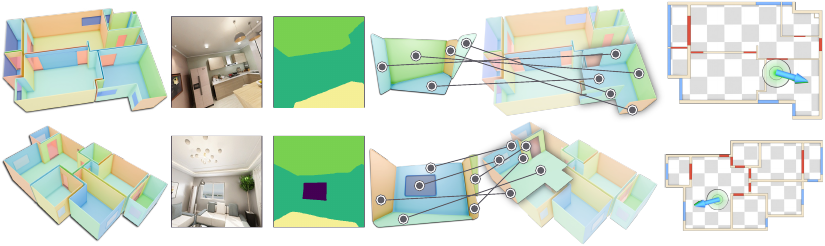}
          };

          {
          \def\subtitley{-0.05}
          \begin{scope}[x={(image.south east)},y={(image.north west)}]
            \node[fill=white, fill opacity=0.7, text opacity=1, inner sep=2pt] at (0.167, \subtitley) {\footnotesize \captiona Input: map \& query};
            \node[fill=white, fill opacity=0.7, text opacity=1, inner sep=2pt] at (0.562, \subtitley) {\footnotesize \captionb Segmentation \& plane recovery \& grounded primitives};
            \node[fill=white, fill opacity=0.7, text opacity=1, inner sep=2pt] at (0.912, \subtitley) {\footnotesize \captionc Localization};
          \end{scope}
          }
        \end{tikzpicture}
        \caption{
          \textbf{Qualitative results on \structured.}
          \captionb We present the predicted layout segmentation, the 3D layout recovered from the query image, and the grounded map primitives, with the ceilings removed for clarity.
          \captionc We show \qfloclegend{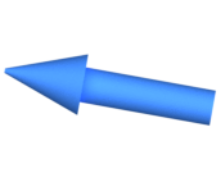}\textcolor{tikzColorMethodOurs}{our localization result} together with \qfloclegend{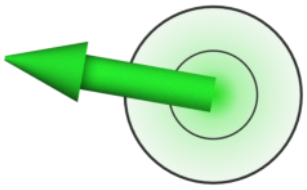}\textcolor{tikzColorMethodGT}{\gtpose}.
        }
        \label{fig:qualitative-s3d-11}
        \vspace{-.7em}
      \end{figure}

      We first present results on \structuredofficial in \cref{tab:exp_s3d_perspective169_door_closed}, where we also highlight several methodological differences:
      \enparen{1} all prior methods adopt the \renderandcompare paradigm, which involves rendering the maps into alternative representations~\enparen{\eg, a set of semantic panoramas} by sweeping the entire candidate pose space;
      \enparen{2} only \spvloc~\cite{gardSPVLocSemanticPanoramic2024} and our method utilize 3D layout models for 6-DoF localization.
      \Cref{tab:exp_s3d_perspective169_door_closed} shows that our method achieves top performance across evaluation metrics, and it is the only approach that \enparen{1} operates in a feed-forward pipeline without requiring exhaustive map rendering, and \enparen{2} does not depend on FoV priors.
      Moreover, our method surpasses \spvloc on the \structuredresampled dataset, as reported in \cref{tab:exp_s3d_perspective11}.
      In \cref{fig:qualitative-s3d-11}, we show two examples on \structuredresampled, illustrating localization results along with additional predicted outputs.

  \subsection{Analysis}

    \paragraph{Multi-task Learning.}
      We conduct ablation studies to evaluate the benefit of predicting multiple targets in our model.
      To reduce computational cost and better assess convergence behavior, models in all ablation experiments are trained for a reduced budget of 75K iterations.
      For the basic components introduced in \cref{sec:model-architecture}, we verify the benefit of training the network to predict multiple localization targets that go beyond the ray map.
      As shown in \cref{tab:ablating-scannet}, training without downstream predictions on patch-primitive assignment and primitive parameters leads to a noticeable drop in localization performance, whereas excluding the prepended camera and map tokens has a milder and mixed effect.

    \paragraph{Layout Adaptations.}
      Next, we ablate the two adaptations for layout models in \cref{sec:model-adaptions}, with results on \structuredresampled-\TFoVNinety\xspace reported in \cref{tab:ablation-s3d-11}.
      Notably, incorporating semantics alone results in a performance drop, which we attribute to slower convergence caused by the fusion of semantics.
      However, the joint integration of semantics and room isolation improves overall performance, indicating a synergistic effect of the two adaptations for this task.

      \begin{table}[htbp]
        \centering
        \vspace{-1em}
        \begin{minipage}[c]{0.56\linewidth}
          \centering
          \captionof{table}{\textbf{Ablation study of each downstream prediction head} in the base model on \scannet.}
          \vspace{.1em}
          \setlength{\tabcolsep}{.5pt} %
          \renewcommand{\arraystretch}{1.25}
          \resizebox{\linewidth}{!}{%
            
\begin{tabular}{lcccccc}
  \Xhline{3\arrayrulewidth}                                                                                                                                                                                                                                                 \\[-1em]
                     & \multicolumn{2}{c}{\TMedianPoseErr} & \multicolumn{3}{c}{\TAccThres} & \multirow{2}{*}{\rotatebox[origin=c]{0}{\makecell[t]{\TAblationMatchFScore}}}                                                                                                 \\[0em]
  \cmidrule(r){2-3} \cmidrule(r){4-6}                                                                                                                                                                                                                                       \\[-1.55em]
  \TMethods          & \TMedianRotErr                      & \TMedianPosErr                 & \TAccFine                                                                     & \TAccMed                      & \TAccCoarse                   &                               \\[0em]
  \Xhline{.3\arrayrulewidth}                                                                                                                                                                                                                                                \\[-1em]
  \TablationBase     & \TRankSecond{\tablenum{3.0}}        & \TRankFirst{\tablenum{0.16}}   & \TRankFirst{\tablenum{62.0}}                                                  & \TRankFirst{\tablenum{92.8}}  & \TRankSecond{\tablenum{95.2}} & \TRankFirst{\tablenum{81.0}}  \\[0em]
  \Xhline{.3\arrayrulewidth}                                                                                                                                                                                                                                                \\[-1em]
  \TablationNoAssign & \TRankThird{\tablenum{3.1}}         & \TRankThird{\tablenum{0.18}}   & \tablenum{54.7}                                                               & \tablenum{87.0}               & \tablenum{92.5}               & \Tna                          \\[0em]
  \TablationNoParam  & \tablenum{3.4}                      & \tablenum{0.19}                & \TRankThird{\tablenum{54.9}}                                                  & \TRankThird{\tablenum{89.6}}  & \TRankThird{\tablenum{95.0}}  & \TRankThird{\tablenum{80.0}}  \\[0em]
  \TablationNoPose   & \TRankFirst{\tablenum{2.9}}         & \TRankSecond{\tablenum{0.17}}  & \TRankSecond{\tablenum{59.7}}                                                 & \TRankSecond{\tablenum{91.0}} & \TRankFirst{\tablenum{95.5}}  & \TRankSecond{\tablenum{80.8}} \\[0em]
  \Xhline{3\arrayrulewidth}                                                                                                                                                                                                                                                 \\[-1em]
\end{tabular}

          }
          \label{tab:ablating-scannet}
        \end{minipage}
        \hfill
        \begin{minipage}[c]{0.40\linewidth}
          \centering
          \captionof{table}{\textbf{Ablating adaptations} for layout localization on \structuredresampled-\TFoVNinety.}
          \vspace{.1em}
          \setlength{\tabcolsep}{2pt} %
          \renewcommand{\arraystretch}{1.25}
          \resizebox{\linewidth}{!}{%
            
\begin{tabular}{lccc}
  \Xhline{3\arrayrulewidth}                                                                                                                                                                  \\[-1em]
                          & \multicolumn{2}{c}{\TfLocAblationPoseMedianError} & \multirow{2}{*}{\rotatebox[origin=c]{0}{\makecell[t]{\TAblationPoseRecall}}}                                 \\[0em]
  \cmidrule(r){2-3}                                                                                                                                                                          \\[-1.55em]
  \TMethods               & \TfLocRotError                                    & \TfLocTransError                                                             &                               \\[0em]
  \Xhline{.3\arrayrulewidth}                                                                                                                                                                 \\[-1em]
  \TablationVanillaPGGT   & \TRankThird{\tablenum{0.69}}                      & \TRankThird{\tablenum{15.2}}                                                 & \tablenum{90.5}               \\[0em]
  \TablationPGGTSemantics & \tablenum{0.88}                                   & \tablenum{17.3}                                                              & \TRankThird{\tablenum{91.6}}  \\[0em]
  \TablationPGGTISOLATION & \TRankSecond{\tablenum{0.53}}                     & \TRankSecond{\tablenum{11.2}}                                                & \TRankSecond{\tablenum{93.8}} \\[0em]
  \Xhline{.3\arrayrulewidth}                                                                                                                                                                 \\[-1em]
  \TablationPGGTFULL      & \TRankFirst{\tablenum{0.46}}                      & \TRankFirst{\tablenum{9.3}}                                                  & \TRankFirst{\tablenum{95.5}}  \\[0em]
  \Xhline{3\arrayrulewidth}                                                                                                                                                                  \\[-1em]
\end{tabular}

          }
          \label{tab:ablation-s3d-11}
        \end{minipage}
        \vspace{-.5em}
      \end{table}

\begin{figure}[htbp]
        \centering
        \begin{minipage}[t]{0.58\linewidth}
          \vspace{0pt}
          \centering
          \setlength{\tabcolsep}{2pt}
          \renewcommand{\arraystretch}{1.25}
          \resizebox{.95\linewidth}{!}{
\begin{tabular}{lccccc}
  \Xhline{3\arrayrulewidth}                                                                                                                                                                \\[-1em]
                    & \multicolumn{2}{c}{\TMedianPoseErr} & \multicolumn{3}{c}{\TAccThres}                                                                                                 \\[0em]
  \cmidrule(r){2-3} \cmidrule(r){4-6}                                                                                                                                                      \\[-1.55em]
  \TMethods         & \TMedianRotErr                      & \TMedianPosErr                 & \TAccFine                     & \TAccMed                      & \TAccCoarse                   \\[0em]
  \Xhline{.3\arrayrulewidth}                                                                                                                                                               \\[-1em]
  \TablationBase    & \TRankFirst{\tablenum{3.0}}         & \TRankFirst{\tablenum{0.16}}   & \TRankFirst{\tablenum{62.0}}  & \TRankFirst{\tablenum{92.8}}  & \TRankSecond{\tablenum{95.2}} \\[0em]
  \Xhline{.3\arrayrulewidth}                                                                                                                                                               \\[-1em]
  \TablationNoPE    & \TRankSecond{\tablenum{3.2}}        & \TRankSecond{\tablenum{0.17}}  & \TRankSecond{\tablenum{59.3}} & \TRankSecond{\tablenum{90.7}} & \TRankFirst{\tablenum{95.4}}  \\[0em]
  \TablationNoParam & \TRankThird{\tablenum{3.4}}         & \TRankThird{\tablenum{0.19}}   & \TRankThird{\tablenum{54.9}}  & \TRankThird{\tablenum{89.6}}  & \TRankThird{\tablenum{95.0}}  \\[0em]
  \TablationNoBoth  & \tablenum{3.4}                      & \tablenum{0.19}                & \tablenum{51.1}               & \tablenum{89.4}               & \tablenum{95.0}               \\[0em]
  \Xhline{3\arrayrulewidth}                                                                                                                                                                \\[-1em]
\end{tabular}
}
          \captionof{table}{\textbf{Ablating plane-specific components} on \scannet.}
          \label{tab:ablation-plane-components}
        \end{minipage}%
        \hfill
        \begin{minipage}[t]{0.41\linewidth}
          \vspace{-6pt}
          \centering
          \includegraphics[width=.95\linewidth]{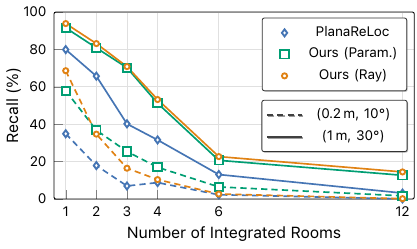}
          \vspace{-3pt}
          \captionof{figure}{\textbf{Quantitative results } on Integrated Rooms~\cite{brachmannExpertSampleConsensus2019}.}
          \label{fig:integrated-rooms}
        \end{minipage}
      \end{figure}

    \paragraph{Explicit Plane Parameters.}
      To explore the potential for extending our framework to non-planar structures, we examine its reliance on explicit plane parameters.
      These parameters enter training through two components: an MLP that provides positional encodings for map primitives and a parameter prediction task supervised by $\lossparameter$.
      We ablate these components individually and jointly on \scannet, with results reported in \cref{tab:ablation-plane-components}.
      These components primarily improve fine-grained localization accuracy, while coarse recall remains largely unchanged even when both are removed.
      This suggests that primitive grounding need not rely on explicit plane parameters, motivating future extensions to non-planar map primitives.

    \paragraph{Evaluation on Multiple Rooms.}
      To assess generalization to larger maps, we follow the standard Integrated Rooms protocol~\cite{brachmannExpertSampleConsensus2019}, arranging \twelvescenes~\cite{valentinLearningNavigateEnergy2016} rooms on a 2D grid with \qty{5}{\meter} cells and progressively merging adjacent rooms.
      We evaluate the original model without fine-tuning, localizing each query within the integrated map.
      As shown in \cref{fig:integrated-rooms}, our method consistently outperforms PlanaReLoc~\cite{yePlanaReLocCameraRelocalization2026} across evaluated scales, although recall decreases substantially as more rooms are integrated, likely due to repetitive structures and the distribution gap from room-scale training.
      These results demonstrate generalization to artificially integrated multi-room maps while highlighting the limitations of full-map inference, motivating coarse-to-fine localization for larger environments.

\section{Conclusion}

  In this paper, we have presented \methodname, a cross-modal transformer network that estimates camera poses from 3D planar primitive-based maps in a feed-forward manner.
  The design of \methodname serves as an instantiation of the new concept that we refer to as \emph{primitive-grounded localization}.
  By jointly reasoning over query image patches and map primitives through two modality-specific interconnected transformer stacks, \methodname predicts multiple localization targets, with poses derived from the densely predicted ray map yielding the best performance.
  Comprehensive evaluation across hundreds of surface maps that comprise dense planar primitives demonstrates the superior performance of our method.
  Furthermore, our method naturally extends to \layoutnameplural, outperforming existing layout localization methods and highlighting the potential of primitive-grounded localization as a promising paradigm for diverse primitive-based map representations.

\section{Acknowledgements and Disclosures}

  This work was supported in part by the Beijing Natural Science Foundation (No.~L223003), the National Natural Science Foundation of China (No.~U22B2055, 62402494, 62273345, 62402495), the Leader-recruiting Bidding Projects for Key Industrial Technology Research of Henan Province (No.~261000320200), the Shandong Provincial Natural Science Foundation (No.~ZR2025LGY009) and the CIE-Tencent Robotics X Rhino-Bird Focused Research Program.
  The authors declare no competing interests.

{

}

\clearpage
\appendix
\section{Appendix}

  \noindent In the appendix, we provide the following:
  \begin{itemize}
    \item Details on data preparation~\enparen{\cref{sec:appendix-dataset}} and a brief summary of dataset statistics~\enparen{\cref{tab:appendix-dataset-statistics}}.
    \item Implementation details on data augmentations, training losses, and plane recovery~\enparen{\cref{sec:appendix-implementation}}.
    \item Ablation of pathway-specific supervision and its comparison with full multi-task supervision~\enparen{\cref{tab:ablation-pathway-supervision}}.
    \item Evaluations on \scannet and \twelvescenes using cumulative accuracy curves to account for all thresholds~\enparen{\cref{fig:cumulative-error}}.
    \item Analysis on how the number of map primitives observed affects localization~\enparen{\cref{fig:dependence-on-plane-num}}.
    \item Additional evaluations on plane recovery~\enparen{\cref{tab:appendix-plane-recovery,fig:per-pixel-and-per-plane-recalls}}.
    \item Additional evaluation results on layout localization~\enparen{\cref{tab:appendix-layout}}.
    \item Analysis on how localization performance varies with parameter estimation quality~\enparen{\cref{fig:depth-correlation}}.
    \item More qualitative results on surface maps~\enparen{\cref{fig:more-qualitative-surface}} and layout models~\enparen{\cref{fig:more-qualitative-layout}}.
    \item Discussions on failure cases~\enparen{\cref{fig:surface-failures,fig:layout-failures}}, limitations, and future work.
  \end{itemize}

\subsection{Datasets}
  \label{sec:appendix-dataset}

  Our model is trained on two independent sets of data, targeting localization on the \mapname and the \layoutname, respectively.
  \Cref{tab:appendix-dataset-statistics} summarizes the statistics of these training datasets, and we provide detailed descriptions in the following.

  \paragraph{\titlecap{\mapname.}}
    For surface maps, we inherit the data processing pipeline from \planareloc~\cite{yePlanaReLocCameraRelocalization2026}, expanding the training set to three public indoor datasets.
    Specifically, maps are extracted from the officially provided dense reconstructions using the sequential RANSAC~\cite{fischlerRandomSampleConsensus1981} plane-fitting scripts from \cite{liuPlaneRCNN3DPlane2019,xiePlanarReconRealtime3D2022}; the extracted maps are further compressed with polygon simplification algorithms~\cite{geoscontributorsGEOSComputationalGeometry2025}.
    Training images of the real-world datasets are sampled from the original RGB-D sequences at regular intervals, and each sampled frame will be checked for adequate primitive observations and precise pose alignment.
    MP3D-Habitat is the only synthetic dataset curated from \matterport by \cite{jinPlanarSurfaceReconstruction2021} whose training images are generated using Habitat~\cite{savvaHabitatPlatformEmbodied2019}.
    The resolution of training images is standardized to \num{640}$\times$\num{480}.
    Finally, per-pixel map primitive IDs are rendered onto training images to obtain the dense pixel-primitive associations, which can further be leveraged to calculate the ground-truth per-pixel parameter map $\parametermap$ or downsampled into the patch-primitive assignment matrix $\assignment$ for supervision.

    \begin{table}[htbp]
      \centering
      \vspace{-.6em}
      \caption{
        \textbf{Statistics of training datasets.}
        Two groups of datasets are used to train the model for localization on surface maps and layout models, respectively.
      }
      \vspace{.1em}
      \setlength{\tabcolsep}{6pt} %
      \renewcommand{\arraystretch}{1.2}
      \resizebox{\linewidth}{!}{%
        
\begin{tabular}{lcccccc}
  \Xhline{3\arrayrulewidth}                                                                                                                                                                                                                                                                                                                                                                                                                                                                                                                             \\[-1em]
                                                                      & \multirow{2}{*}{\rotatebox[origin=c]{0}{\makecell[t]{\TDatasetMapType}}} & \multirow{2}{*}{\rotatebox[origin=c]{0}{\makecell[t]{\TDatasetDomain}}} & \multirow{2}{*}{\rotatebox[origin=c]{0}{\makecell[t]{\TDatasetCameraType}}} & \multirow{2}{*}{\rotatebox[origin=c]{0}{\makecell[t]{\TDatasetNumFrames}}} & \multirow{2}{*}{\rotatebox[origin=c]{0}{\makecell[t]{\TDatasetNumScenes}}} & \multirow{2}{*}{\rotatebox[origin=c]{0}{\makecell[t]{\TDatasetAvgFramesPerScene}}} \\[0em]
  \\[-1.55em]
  \TDatasetName                                                       &                                                                          &                                                                         &                                                                             &                                                                            &                                                                            &                                                                                    \\[0em]
  \Xhline{.3\arrayrulewidth}                                                                                                                                                                                                                                                                                                                                                                                                                                                                                                                            \\[-1em]
  \scannet~\cite{daiScanNetRichlyannotated3D2017}                     & \multirow{4}{*}{\rotatebox[origin=t]{0}{\makecell[t]{\DetailedMap}}}     & \DatasetRealWorld                                                       & \PerspectiveCamera                                                          & \num{98533}                                                                & \num{1210}                                                                 & \num{81}                                                                           \\[0em]
  \scannetpp~\cite{yeshwanthScanNetHighfidelityDataset2023}           &                                                                          & \DatasetRealWorld                                                       & \PerspectiveCamera                                                          & \num{216645}                                                               & \num{856}                                                                  & \num{253}                                                                          \\[0em]
  \matterport~\cite{changMatterport3DLearningRGBD2017}                &                                                                          & \DatasetRealWorld                                                       & \PerspectiveCamera                                                          & \num{54578}                                                                & \num{2195}                                                                 & \num{25}                                                                           \\[0em]
  \matterportsync~\cite{savvaHabitatPlatformEmbodied2019}             &                                                                          & \DatasetSynthetic                                                       & \PerspectiveCamera                                                          & \num{24568}                                                                & \num{2195}                                                                 & \num{11}                                                                           \\[0em]
  \Xhline{.3\arrayrulewidth}                                                                                                                                                                                                                                                                                                                                                                                                                                                                                                                            \\[-1em]
  \structuredofficial~\cite{zhengStructured3DLargePhotorealistic2020} & \multirow{2}{*}{\rotatebox[origin=t]{0}{\makecell[t]{\MinimalistMap}}}   & \DatasetSynthetic                                                       & \PerspectiveCamera                                                          & \num{148341}                                                               & \num{3250}                                                                 & \num{46}                                                                           \\[0em]
  \structuredresampled~\cite{gardSPVLocSemanticPanoramic2024}         &                                                                          & \DatasetSynthetic                                                       & \PanoramicCamera                                                            & \num{21835}                                                                & \num{3250}                                                                 & \num{7}                                                                            \\[0em]
  \Xhline{3\arrayrulewidth}                                                                                                                                                                                                                                                                                                                                                                                                                                                                                                                             \\[-1em]
\end{tabular}

      }
      \label{tab:appendix-dataset-statistics}
      \vspace{-.4em}
    \end{table}

  \paragraph{\titlecap{\layoutname.}} \structuredofficial and \structuredresampled share the same collection of layout models, but differ in the query images used for training and evaluation.
    \structuredofficial comprises officially provided high-resolution perspective images in a format of \num{16}:\num{9} with a fixed FoV of \qty{80}{\degree}, which are resized to the resolution of \num{640}$\times$\num{360} in our experiments.
    On the other hand, the \structuredresampled dataset proposed by ~\cite{gardSPVLocSemanticPanoramic2024} starts from the original panorama images and samples training perspective images on the fly with random camera orientations~\enparen{yaw $\in$ [-\qty{180}{\degree}, \qty{180}{\degree}], pitch $\in $ [-\qty{10}{\degree}, \qty{10}{\degree}] and roll $\in$ [-\qty{10}{\degree}, \qty{10}{\degree}]} as well as random FoVs from \qty{45}{\degree} to \qty{135}{\degree}.
    These perspective images are sampled at a resolution of \num{320}$\times$\num{320}.
    The sampled images will be verified for observations of at least three semantic classes~\enparen{\eg, wall-ceiling-floor and wall-window-floor} and adequate coverage of the room interior~\enparen{\ie, doors and windows not occupying more than \qty{70}{\percent} of the image}.
    For efficiency, we generate at least \num{100} verified candidates of yaw, pitch, roll and FoV for each panorama when preparing the dataset, and during training, images are directly sampled from these candidates, avoiding the redundant computation for verification.
    Likewise, annotations of per-pixel primitive IDs are rendered onto the original panorama, and then can be sampled to generate perspective images for training on the fly.

  \subsection{Implementation Details}
  \label{sec:appendix-implementation}
  \paragraph{Input Augmentations.}
    During training, we apply brightness, color and contrast jittering, and randomly rescale the query images to heights between \num{384} and \num{600} pixels before cropping or padding them to the final resolution of \num{640}$\times$\num{480}.
    The maps are normalized into a unit box and randomly rescaled and rotated by a random yaw angle, with map primitives shuffled before being fed into the model.
    For samples from \matterport, maps are assembled online from submaps of multiple regions according to the query images.
    Specifically, we merge the submaps corresponding to regions observed by the query image together with those of spatially nearby regions, resulting in dynamic map coverage.

  \paragraph{Training Losses.}
    We elaborate on the loss terms introduced in \cref{equ:losses}.
    Given the ground-truth patch-primitive assignment matrix $\assignment$, we denote the set of positive patch-primitive pairs as $\assignmentPos = \{(i, j) | \assignment_{ij} = 1\}$ and unassignable patch and primitive sets as $\unassignablePatch = \{i | \sum_j \assignment_{ij} = 0\}$ and $\unassignablePrimitive = \{j | \sum_i \assignment_{ij} = 0\}$, respectively.
    Then, the assignment loss is defined as the negative log-likelihood of the soft assignment matrix $\softassignment$ and unassignable predictions:
    \begin{equation}
      \lossassignment = - \bigg(
      \frac{1}{|\assignmentPos|}\sum_{(i,j)\in \assignmentPos} \log \softassignment_{ij} + \frac{1}{2|\unassignablePatch|}\sum_{i\in \unassignablePatch} \log ( 1 - \descriptorconfidence^q_i) + \frac{1}{2|\unassignablePrimitive|}\sum_{j\in \unassignablePrimitive} \log (1 - \descriptorconfidence^m_j)\bigg).
      \label{equ:loss}
    \end{equation}
    For geometric supervision on primitive parameters, $\lossparameter$ is the combination of the $\normlone$ loss and the cosine similarity loss $\losscos$, computed on the per-pixel parameter prediction $\hat{\parametermap}^q\in\R^{3\times H\times W}$ of the query image as well as the per-primitive prediction $\hat{\parametermap}^m\in\R^{3\times \mapplanenum}$ of the map:
    \begin{equation}
      \lossparameter = \sum_{r \in \{q,m\}}\left(\normlone(\hat{\parametermap}^r, \parametermap^r) + \losscos(\hat{\parametermap}^r, \parametermap^r)\right).
    \end{equation}
    The ray map loss $\lossray$ and the pose decoding loss $\losspose$ are both based on the $\normlone$ norm:
    \begin{align}
      \lossray  & = \raymapuncertainty\odot \normlone(\hat{\raymap}, \raymap) - \lambda_u \log\raymapuncertainty, \\
      \losspose & = \normlone(\hat{\querypose}^q, \querypose^q) + \normlone(\hat{\querypose}^m, \querypose^m).
    \end{align}
    Here, $\raymapuncertainty$ is the ray uncertainty map and $\odot$ is the channel-broadcast element-wise product.
    The target pose $\querypose^m$ for the map token is set to the identity pose with the rotation quaternion $\canonicalposequaternion=[0, 0, 0, 1]$ and translation $\canonicalposetranslation=[0, 0, 0]$.

  \paragraph{Plane Recovery.}
    Recovering piecewise planar primitives from the query image is not the main focus of this work, since camera poses can be directly derived from the predicted ray map.
    Nevertheless, plane recovery is a geometric by-product of our method, and we briefly describe how its results are obtained for visualizations~\enparen{as shown in \cref{fig:comparison_with_planareloc,fig:qualitative-s3d-11}} and the subsequent analysis in \cref{sec:appendix-analysis}.
    The task of plane recovery requires both 2D plane segments and their corresponding plane parameters.
    The 2D plane segments are first obtained by partitioning the predicted parameter map $\hat{\parametermap}^q \in \mathbb{R}^{3 \times H \times W}$ according to the predicted patch-primitive assignment matrix $\hat{\assignment}$.
    Since the network does not enforce local consistency, \ie, densely predicted per-pixel parameters within each plane segment are not guaranteed to be consistent, the final plane parameters for each segment are robustly estimated with RANSAC.

\subsection{Additional Evaluation and Analysis}
  \label{sec:appendix-analysis}

  \paragraph{Pathway-specific Supervision.}
    The ablation in \cref{tab:ablating-scannet} evaluates auxiliary prediction tasks while keeping Ray-based inference fixed; removing $\lossray$ and switching inference pathways would change both supervision and inference.
    To assess each pathway separately, we train models using only the supervision required by their respective inference pathways and compare them with the full model on \scannet in \cref{tab:ablation-pathway-supervision}.
    All three pathways independently enable localization, with Ray-based inference performing best overall among the individually supervised variants.
    Full multi-task supervision further improves Ray-based localization, demonstrating the complementary benefits of the prediction tasks.

    \begin{table}[htbp]
      \centering
      \caption{\textbf{Pathway-specific supervision} on \scannet.
        Each pathway is trained with its required supervision only; Full (Ray) uses all supervision terms.
      }
      \label{tab:ablation-pathway-supervision}
      \setlength{\tabcolsep}{3pt}
      \renewcommand{\arraystretch}{1.2}
      \resizebox{\linewidth}{!}{%
        
\begin{tabular}{llccccc}
  \toprule
               &                                                     & \multicolumn{2}{c}{\TMedianPoseErr} & \multicolumn{3}{c}{\TAccThres}                                                 \\
  \cmidrule(lr){3-4} \cmidrule(lr){5-7}
  Pathway      & Supervision                                         & \TMedianRotErr                      & \TMedianPosErr                 & \TAccFine     & \TAccMed      & \TAccCoarse   \\
  \midrule
  Param.-based & $\lossassignment+\lossparameter$                    & 3.1                                 & 0.23                           & 44.2          & 70.6          & 79.5          \\
  Token-based  & $\losspose$                                         & 4.4                                 & 0.28                           & 39.9          & 83.7          & 87.2          \\
  Ray-based    & $\lossray$                                          & 3.4                                 & 0.20                           & 51.7          & 87.6          & 90.9          \\
  \midrule
  Full (Ray)   & $\lossassignment+\lossparameter+\losspose+\lossray$ & \textbf{3.0}                        & \textbf{0.16}                  & \textbf{62.0} & \textbf{92.8} & \textbf{95.2} \\
  \bottomrule
\end{tabular}

      }
    \end{table}

\begin{figure}[htbp]
      \centering
      \includegraphics{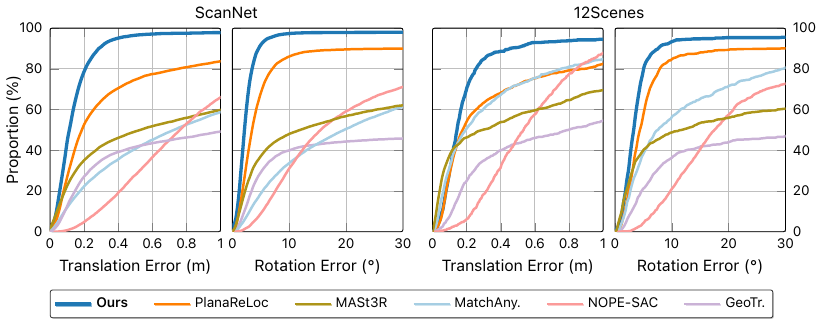}
      \caption{\textbf{Cumulative accuracy curves} on surface maps.
        We plot the proportion of successfully localized frames as a function of varying translation and rotation error thresholds.
      }
      \label{fig:cumulative-error}
    \end{figure}

  \paragraph{Cumulative Accuracy Curves.}
    As shown in \cref{fig:cumulative-error}, the cumulative accuracy curves illustrate the proportion of camera poses whose translation and rotation errors fall below varying thresholds from \qtyrange[range-phrase=\ to\ ]{0}{1}{\meter} and from \qtyrange[range-phrase=\ to\ ]{0}{30}{\degree}.
    Serving as a more detailed performance comparison, the curves on both the ScanNet and 12Scenes datasets highlight the superior robustness and accuracy of our method across both translation and rotation.

    \begin{figure*}[htbp]
      \begin{subfigure}[b]{0.33\textwidth}
        \centering
        \includegraphics{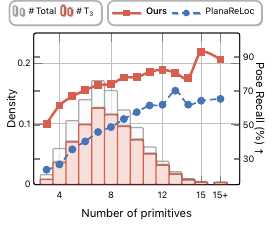}
        \caption{Pose recall.}
        \label{fig:dependence-on-plane-num-a}
      \end{subfigure}%
      \begin{subfigure}[b]{0.33\textwidth}
        \centering
        \includegraphics{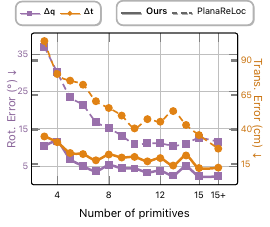}
        \caption{Average pose error.}
        \label{fig:dependence-on-plane-num-b}
      \end{subfigure}%
      \begin{subfigure}[b]{0.33\textwidth}
        \centering
        \includegraphics{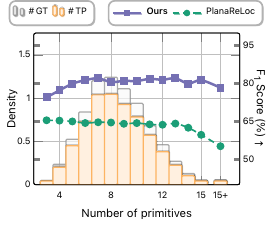}
        \caption{Matching performance.}
        \label{fig:dependence-on-plane-num-c}
      \end{subfigure}%
      \newcommand{\captionleft}{\uline{\emph{Left:}} }
      \newcommand{\captionright}{\uline{\emph{Right:}} }
      \caption{
        \textbf{Analysis of the impact of primitive richness} on \scannet.
        \captiona \captionleft Distribution of all test samples and successfully localized samples~\enparen{by our method} at \AccFine\xspace thresholds \wrt the number of actually observed primitives.
        \captionright Pose recall at \AccFine\xspace versus the number of actually observed primitives.
        \captionb Average pose error curves, where pose errors are averaged over all samples within each bin.
        \captionc \captionleft  Distribution of actually observed map primitives and those successfully matched~\enparen{by our method}.
        \captionright Curves of the matching F$_\text{1}$ score.
      }
      \label{fig:dependence-on-plane-num}
      \vspace{-.8em}
    \end{figure*}

  \paragraph{Impact of Primitive Richness.}
    In practice, camera localization becomes unreliable when query observations are ambiguous and uninformative, \eg, when only a few map primitives are observed.
    To analyze the impact of primitive richness on localization performance, we plot the localization performance as a function of the number of actually observed primitives in \cref{fig:dependence-on-plane-num}.
    As shown in \cref{fig:dependence-on-plane-num}\refcaptiona and \refcaptionb, observing more primitives generally improves localization accuracy.
    Notably, compared to the feature-matching method \planareloc, which depends on explicit primitive correspondences for geometric pose estimation and can become unstable when few primitives are observed, our method maintains robust performance, yielding reliable camera poses even under sparse observations.
    Moreover, as shown in \cref{fig:dependence-on-plane-num}\refcaptionc, when the query observation becomes cluttered with more primitives, there is a notable drop in the matching F\textsubscript{1} score of \planareloc due to its degraded plane recovery performance as planes in richer observations tend to be less salient and harder to recover.
    In contrast, our method performs exhaustive interaction between image patches and map primitives during the single feed-forward pass, allowing the model to finely ground the query observations to the map primitives, and thereby exhibiting a more stable matching performance.

\begin{figure}[t]
      \centering
      \includegraphics{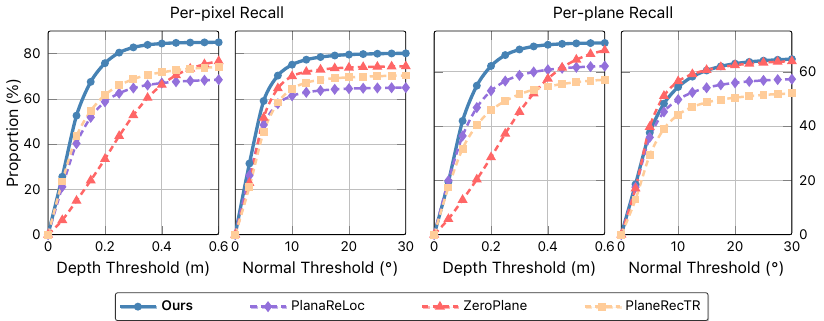}
      \caption{\textbf{Per-pixel and per-plane recalls} on our \scannet test set, which are comprehensive metrics for evaluating the plane recovery quality.}
      \label{fig:per-pixel-and-per-plane-recalls}
    \end{figure}

  \paragraph{Evaluation on Plane Recovery.}
    While poses can be derived directly from the predicted ray maps and plane recovery is thus not the primary focus of this work, we also use our plane recovery results to evaluate the predictions on primitive parameters enabled by the corresponding head in our model, to assess the quality of geometry learning.
    Following the standard metrics in the literature~\cite{yangRecovering3DPlanes2018,liuPlaneRCNN3DPlane2019}, the performance of plane recovery is evaluated from two perspectives: the segmentation quality of 2D plane instances and the accuracy of the depth estimation from the predicted plane parameters.
    We compare our method with two end-to-end monocular plane recovery methods~\cite{shiPlaneRecTRUnifiedQuery2023,liuInthewild3DPlane2025}, as well as the method used in \planareloc on our \scannet test set, and the results are summarized in \cref{tab:appendix-plane-recovery}.
    A more comprehensive evaluation using per-pixel and per-plane recalls~\cite{liuPlaneNetPiecewisePlanar2018} is shown in \cref{fig:per-pixel-and-per-plane-recalls}.
    Although not trained with explicit supervision on plane recovery, by reasoning jointly over map primitives, our method significantly outperforms the monocular methods.

    \begin{table}[t]
      \centering
      \vspace{-.5em}
      \caption{
        \textbf{Comparison of plane recovery results} on our \scannet test set.
        Although existing monocular methods are not direct competitors, we quantitatively compare their plane recovery quality with ours using standard segmentation and depth estimation metrics.
        The {\setlength{\fboxsep}{1pt}\colorbox{tableRankFirst}{\textbf{best}}} and {\setlength{\fboxsep}{1pt}\colorbox{tableRankSecond}{\emph{second}}} are highlighted.
      }
      \vspace{.1em}
      {
        \renewcommand{\TRankThird}[1]{#1}
        \setlength{\tabcolsep}{6pt} %
        \renewcommand{\arraystretch}{1.2}
        \resizebox{\linewidth}{!}{%
          
\begin{tabular}{lccccccccc}
  \Xhline{3\arrayrulewidth}                                                                                                                                                                                                                                                                                                                                                      \\[-1em]
                                                    & \multicolumn{3}{c}{\TPlaneRecoverySegmentation} & \multicolumn{6}{c}{\TPlaneRecoveryDepth}                                                                                                                                                                                                                                 \\[0em]
  \cmidrule(r){2-4} \cmidrule(r){5-10}                                                                                                                                                                                                                                                                                                                                           \\[-1.55em]
  \TMethods                                         & \TSegmentationRI                                & \TSegmentationVOI                        & \TSegmentationSC              & \TDepthRMSE                   & \TDepthLog                    & \TDepthRel                    & \TDepthAccFine                & \TDepthAccMed                 & \TDepthAccCoarse              \\[0em]
  \Xhline{.3\arrayrulewidth}                                                                                                                                                                                                                                                                                                                                                     \\[-1em]
  \Tplanerectr~\cite{shiPlaneRecTRUnifiedQuery2026} & \TRankThird{\tablenum{0.91}}                    & \TRankThird{\tablenum{0.93}}             & \TRankThird{\tablenum{0.78}}  & \TRankSecond{\tablenum{0.24}} & \TRankSecond{\tablenum{0.05}} & \TRankSecond{\tablenum{0.09}} & \TRankSecond{\tablenum{90.8}} & \TRankSecond{\tablenum{97.7}} & \TRankSecond{\tablenum{99.2}} \\[0em]
  \Tzeroplane~\cite{liuInthewild3DPlane2025}        & \TRankSecond{\tablenum{0.92}}                   & \TRankSecond{\tablenum{0.90}}            & \TRankSecond{\tablenum{0.80}} & \TRankThird{\tablenum{0.34}}  & \TRankThird{\tablenum{0.09}}  & \TRankThird{\tablenum{0.18}}  & \tablenum{65.9}               & \TRankThird{\tablenum{92.1}}  & \TRankThird{\tablenum{97.7}}  \\[0em]
  \TMoGeRANSAC                                      & \tablenum{0.89}                                 & \tablenum{1.28}                          & \tablenum{0.71}               & \tablenum{0.65}               & \tablenum{0.58}               & \tablenum{0.20}               & \TRankThird{\tablenum{82.3}}  & \tablenum{86.0}               & \tablenum{86.5}               \\[0em]
  \Xhline{.3\arrayrulewidth}                                                                                                                                                                                                                                                                                                                                                     \\[-1em]
  \Tours                                            & \TRankFirst{\tablenum{0.94}}                    & \TRankFirst{\tablenum{0.78}}             & \TRankFirst{\tablenum{0.86}}  & \TRankFirst{\tablenum{0.16}}  & \TRankFirst{\tablenum{0.03}}  & \TRankFirst{\tablenum{0.07}}  & \TRankFirst{\tablenum{96.0}}  & \TRankFirst{\tablenum{99.1}}  & \TRankFirst{\tablenum{99.7}}  \\[0em]
  \Xhline{3\arrayrulewidth}                                                                                                                                                                                                                                                                                                                                                      \\[-1em]
\end{tabular}

        }
      }
      \label{tab:appendix-plane-recovery}
      \vspace{-.5em}
    \end{table}

  \paragraph{Additional Evaluation for Layout Localization.}
    Since existing layout-based localization methods mostly operate in 2D, we primarily report 2D localization errors in the main text for comparison.
    As our method performs 6-DoF camera localization using a 3D map, we report the 3D median localization errors in \cref{tab:appendix-layout}.
    Furthermore, we evaluate the predicted layout geometry with depth estimation metrics and assess segmentation results using the standard mIoU and mACC metrics.

    \begin{table}[htbp]
      \centering
      \vspace{-.5em}
      \caption{
        \textbf{Additional Evaluation for Layout Localization.}
      }
      \vspace{.1em}
      {
        \renewcommand{\TRankFirst}[1]{#1}
        \renewcommand{\TRankSecond}[1]{#1}
        \renewcommand{\TRankThird}[1]{#1}
        \setlength{\tabcolsep}{3pt} %
        \renewcommand{\arraystretch}{1.2}
        \resizebox{\linewidth}{!}{%
          
\begin{tabular}{lcccccccccc}
  \Xhline{3\arrayrulewidth}                                                                                                                                                                                                                                                                                                                                                                         \\[-1em]
                            & \multicolumn{2}{c}{\TfLocPoseMedianErrorSixDoF} & \multicolumn{6}{c}{\TPlaneRecoveryDepth} & \multicolumn{2}{c}{\TLayoutSegmentation}                                                                                                                                                                                                                                 \\[0em]
  \cmidrule(r){2-3} \cmidrule(r){4-9} \cmidrule(r){10-11}                                                                                                                                                                                                                                                                                                                                           \\[-1.55em]
  Dataset                   & \TMedianRotErr                                  & \TMedianPosErrCM                         & \TDepthRMSE                              & \TDepthLog                    & \TDepthRel                    & \TDepthAccFine                & \TDepthAccMed                 & \TDepthAccCoarse              & \TSegmentationmIoU            & \TSegmentationmACC            \\[0em]
  \Xhline{.3\arrayrulewidth}                                                                                                                                                                                                                                                                                                                                                                        \\[-1em]
  \TSuppstructuredofficial  & \TRankFirst{\tablenum{0.75}}                    & \TRankSecond{\tablenum{13.0}}            & \TRankSecond{\tablenum{0.18}}            & \TRankSecond{\tablenum{0.03}} & \TRankSecond{\tablenum{0.08}} & \TRankSecond{\tablenum{96.1}} & \TRankSecond{\tablenum{98.6}} & \TRankSecond{\tablenum{99.1}} & \TRankFirst{\tablenum{81.6}}  & \TRankFirst{\tablenum{84.9}}  \\[0em]
  \Xhline{.3\arrayrulewidth}                                                                                                                                                                                                                                                                                                                                                                        \\[-1em]
  \TSuppstructuredresampled & \TRankSecond{\tablenum{0.81}}                   & \TRankFirst{\tablenum{11.1}}             & \TRankFirst{\tablenum{0.11}}             & \TRankFirst{\tablenum{0.02}}  & \TRankFirst{\tablenum{0.05}}  & \TRankFirst{\tablenum{98.3}}  & \TRankFirst{\tablenum{99.6}}  & \TRankFirst{\tablenum{99.8}}  & \TRankSecond{\tablenum{78.3}} & \TRankSecond{\tablenum{80.1}} \\[0em]
  \Xhline{3\arrayrulewidth}                                                                                                                                                                                                                                                                                                                                                                         \\[-1em]
\end{tabular}

        }
      }
      \label{tab:appendix-layout}
      \vspace{-.5em}
    \end{table}

\begin{figure}
      \begin{subfigure}[b]{0.48\linewidth}
        \centering
        \includegraphics{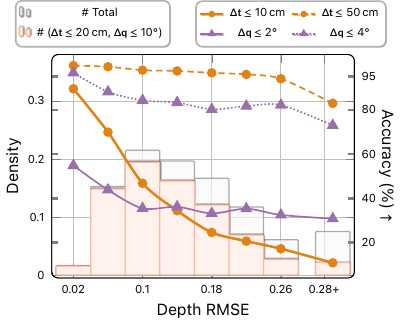}
        \caption{\scannet.}
      \end{subfigure}
      \begin{subfigure}[b]{0.48\linewidth}
        \centering
        \includegraphics{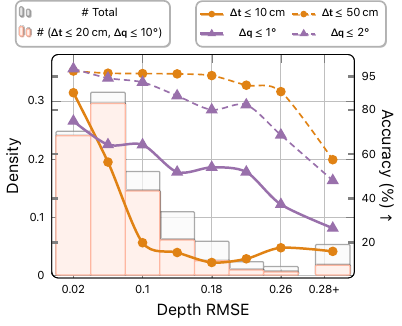}
        \caption{\structuredresampled-\TFoVNinety.}
      \end{subfigure}
      \caption{
        \textbf{Performance trends of depth and pose estimation} on two datasets.
        We plot the depth RMSE distribution as histograms~\enparen{left y-axis} and corresponding pose accuracy curves~\enparen{right y-axis}.
      }
      \label{fig:depth-correlation}
      \vspace{-.5em}
    \end{figure}

  \paragraph{Relationship between Parameter and Pose Estimation.}
    We are interested in finding out how the pose estimation accuracy varies with parameter estimation quality, as in \planareloc, the pose estimation is based on the predicted plane parameters and can be significantly affected by the parameter estimation quality.
    Therefore, we plot the distribution of depth RMSE together with the corresponding pose accuracy curves on both the \scannet and \structuredresampled datasets in \cref{fig:depth-correlation}.
    It can also be observed that for our method, the pose accuracy generally drops as depth RMSE increases.
    However, we note that accurate depth estimation does not necessarily indicate successful localization~\enparen{\eg, due to ambiguous repetitive patterns}, as discussed later in \cref{sec:appendix-limitations}.

\begin{figure}
      \centering
      \begin{tikzpicture}
        \node[anchor=south west,inner sep=0] (image) at (0,0) {
          \includegraphics{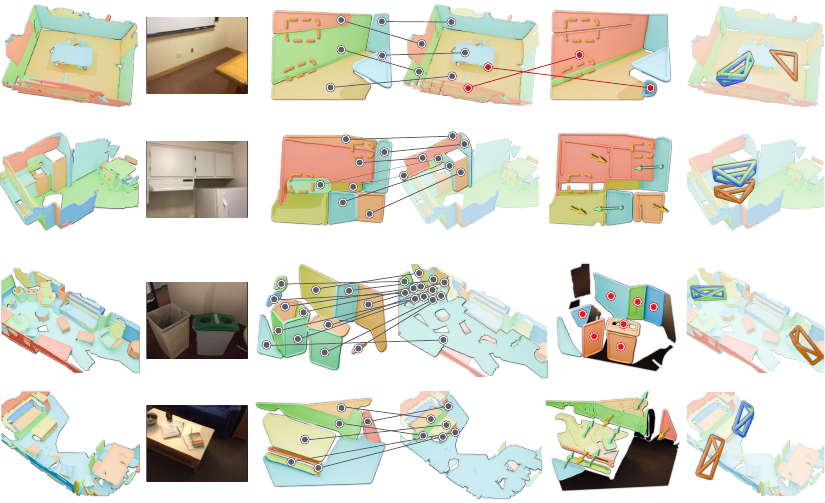}
        };

        {
        \def\subtitley{-0.06}
        \begin{scope}[x={(image.south east)},y={(image.north west)}]
          \node[fill=white, fill opacity=0.7, text opacity=1, inner sep=2pt] at (0.167, \subtitley) {\footnotesize \captiona Input: map \& query};
          \node[fill=white, fill opacity=0.7, text opacity=1, inner sep=2pt] at (0.483, \subtitley) {\footnotesize \captionb Plane recovery \& grounded primitives};
          \node[fill=white, fill opacity=0.7, text opacity=1, inner sep=2pt] at (0.743, \subtitley) {\footnotesize \captionc PlanaReLoc};
          \node[fill=white, fill opacity=0.7, text opacity=1, inner sep=2pt] at (0.92, \subtitley) {\footnotesize \captiond Localization};
        \end{scope}
        }
      \end{tikzpicture}
      \caption{
        \textbf{Qualitative results on surface maps.}
        The first two examples are from \scannet and the last two examples are from \twelvescenes.
        Our method robustly produces accurate pose estimation, with \captionb superior plane recovery and primitive correspondences as by-products,
        while \captionc \planareloc performs suboptimally or even fails due to poor plane recovery, matching outliers and inadequate pose estimation constraints.
        \captiond Comparison of localization results.
        Legend:
        \qplanareloclegend{gt}\textcolor{tikzColorMethodGT}{\gtpose},
        \qplanareloclegend{ours}\textcolor{tikzColorMethodOurs}{\methodname\enparen{Ours}}, and
        \qplanareloclegend{planareloc}\textcolor{tikzColorMethodPlanaReLoc!
          80!black}{\planareloc}.
      }
      \label{fig:more-qualitative-surface}
    \end{figure}

\begin{figure}
      \centering
      \begin{tikzpicture}
        \node[anchor=south west,inner sep=0] (image) at (0,0) {
          \includegraphics{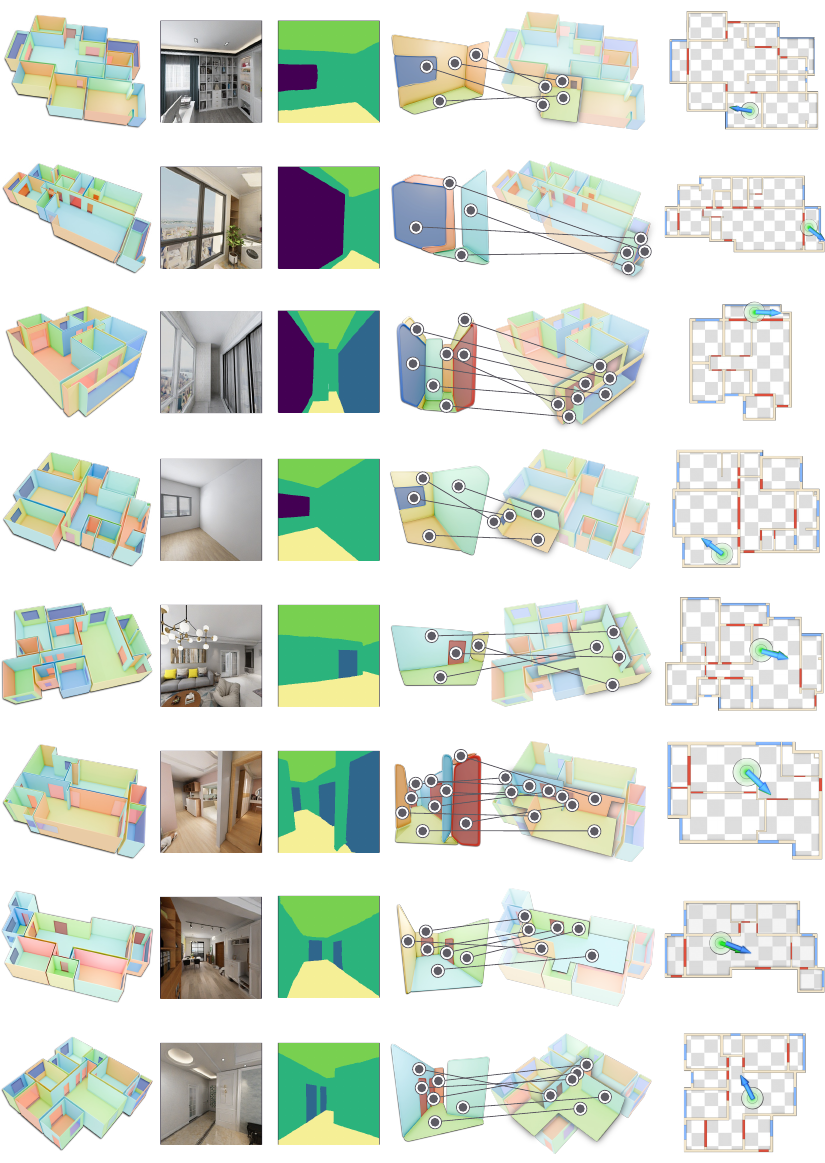}
        };

        {
        \def\subtitley{-0.01}
        \begin{scope}[x={(image.south east)},y={(image.north west)}]
          \node[fill=white, fill opacity=0.7, text opacity=1, inner sep=2pt] at (0.167, \subtitley) {\footnotesize \captiona Input: map \& query};
          \node[fill=white, fill opacity=0.7, text opacity=1, inner sep=2pt] at (0.562, \subtitley) {\footnotesize \captionb Segmentation \& plane recovery \& grounded primitives};
          \node[fill=white, fill opacity=0.7, text opacity=1, inner sep=2pt] at (0.908, \subtitley) {\footnotesize \captionc Localization};
        \end{scope}
        }
      \end{tikzpicture}
      \caption{
        \textbf{Qualitative results on layout models.}
        \captionb We present the predicted layout segmentation, the layout geometry recovered from the query image, and the grounded map primitives, with ceilings removed for clarity.
        \captionc We show \qfloclegend{ours}\textcolor{tikzColorMethodOurs}{the localization result} as well as \qfloclegend{gt}\textcolor{tikzColorMethodGT}{\gtpose}.
      }
      \label{fig:more-qualitative-layout}
    \end{figure}

  \paragraph{Additional Qualitative Results.}
    More visualizations of surface map localization results as well as layout localization results are provided in \cref{fig:more-qualitative-surface} and \cref{fig:more-qualitative-layout}, respectively.
    All results in this work are visualized using Blender~\cite{blenderonlinecommunityBlender3DModelling}.

  \subsection{Limitations and Future Work}
  \label{sec:appendix-limitations}

\begin{figure}
      \centering
      \includegraphics{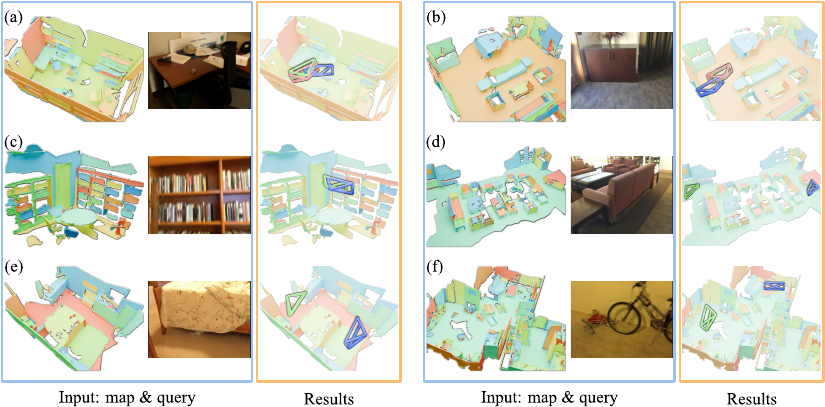}
      \caption{
        \textbf{Representative failure cases on surface maps.}
        \enparen{a\textendash b}
        Cases where the \qplanareloclegend{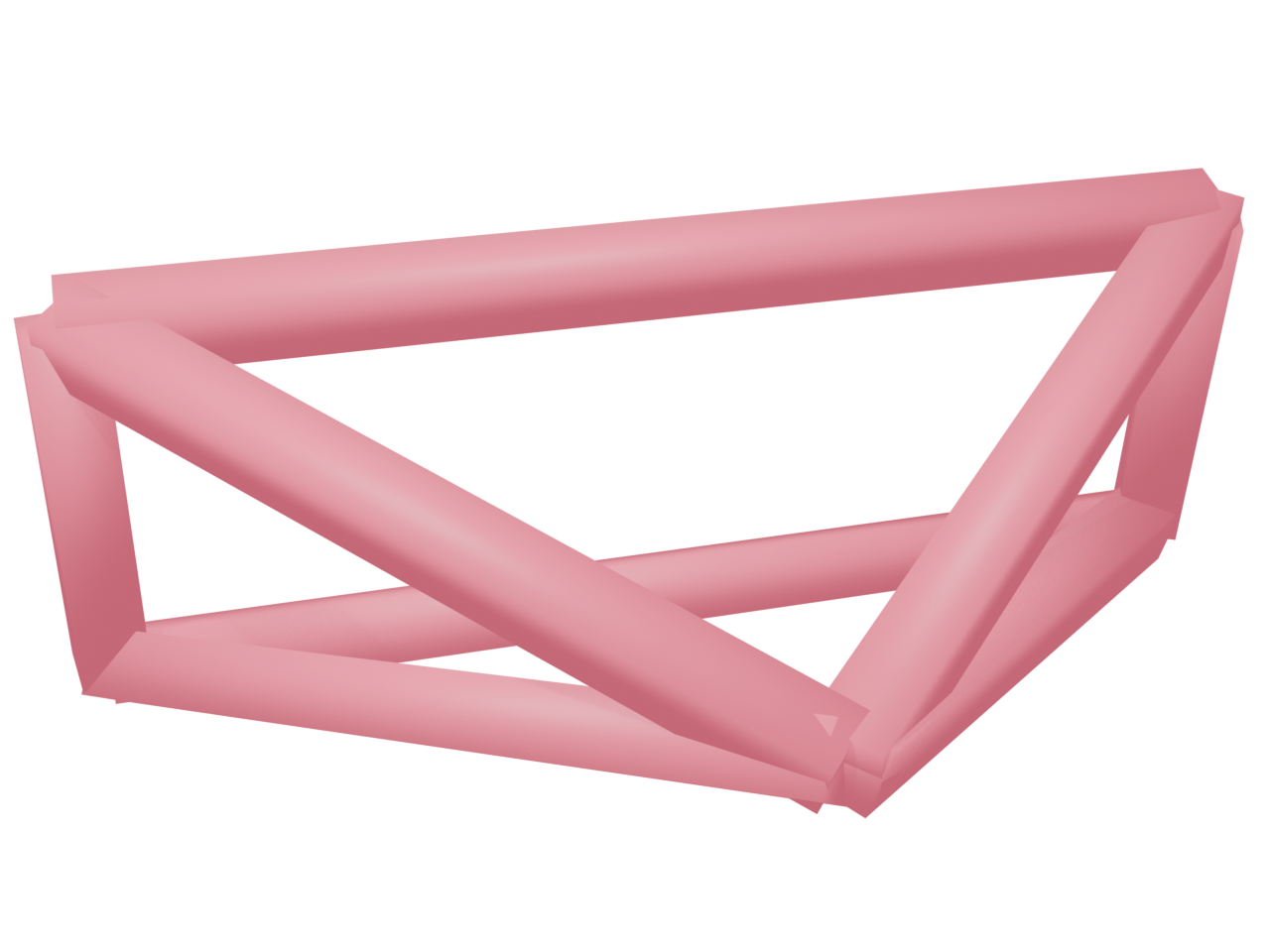}\textcolor{tikzColorMethodOursParam}{\param-based pose} is more accurate than the \qplanareloclegend{ours}\textcolor{tikzColorMethodOurs}{Ray-based pose}.
        \captionc Pose shift under blurry observations.
        \enparen{d\textendash f}
        Failure cases occurring in scenes with limited structural detail or highly repetitive patterns, or caused by uninformative query observations.
      }
      \label{fig:surface-failures}
    \end{figure}

\begin{figure}
      \centering
      \includegraphics{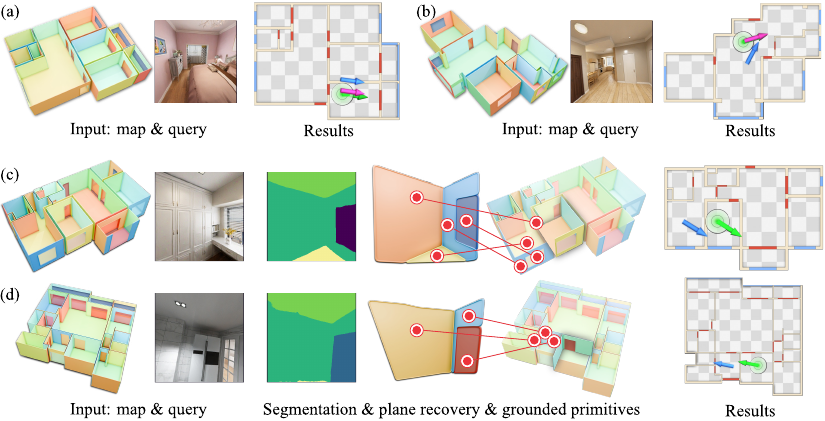}
      \caption{
        \textbf{Representative failure cases on layout models.}
        \enparen{a\textendash b}
        Cases where the \qfloclegend{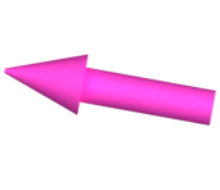}\textcolor{tikzColorMethodOursParam}{\param-based pose} is more accurate than the \qfloclegend{ours}\textcolor{tikzColorMethodOurs}{Ray-based pose}.
        \enparen{c\textendash d}
        Failure cases due to structural ambiguity.
      }
      \label{fig:layout-failures}
    \end{figure}

  \paragraph{Failure Cases.}
    We show representative failure cases of our method on both surface maps and layout models in \cref{fig:surface-failures} and \cref{fig:layout-failures}, respectively.
    In some cases~\enparen{\cref{fig:surface-failures}\refcaptiona, \refcaptionb, and \cref{fig:layout-failures}\refcaptiona, \refcaptionb}, poses derived from the ray map are less accurate than those estimated using predicted primitive parameters.
    We also observe that the failure cases mainly arise from three aspects: \enparen{1} challenging visual conditions, such as motion blur and low lighting; \enparen{2} limited structural details or highly repetitive patterns in the scene; and \enparen{3} uninformative query observations.

  \paragraph{Limitations.}
    First, although our primitive-grounded localization paradigm naturally extends from \mapnameplural to \layoutnameplural, the network is trained separately for each task.
    Then, our work is currently limited to maps that are assembled from planar primitives, which assume the presence of planar surfaces and regularized layouts, and thus may not be suitable for unstructured environments or scenes exhibiting complex or non-planar structures.
    Meanwhile, our model is trained primarily on room-scale indoor environments.
    Although it outperforms PlanaReLoc on artificially integrated multi-room maps without fine-tuning~\enparen{\cref{fig:integrated-rooms}}, recall decreases substantially as map size grows, highlighting the limitations of full-map inference.
    These experiments do not establish performance in real large-scale buildings or outdoor environments, where the distribution of map primitives may differ significantly.

  \paragraph{Future Work.}
    The proposed primitive-grounded localization paradigm can potentially extend to a variety of high-level map representations in the future, such as LoD city models and structured language commands~\cite{avetisyanSceneScriptReconstructingScenes2024}.
    This is enabled by the flexibility and scalability of our feed-forward localization framework, which allows it to accommodate diverse map representations and even unify localization across different map modalities within a single model.
    In addition, extending the method to panoramic or sequential inputs offers another promising direction for practical application, as richer visual cues and temporal context could effectively help resolve structural ambiguities.

  \subsection{Societal Impacts}
  \label{sec:appendix-societal-impacts}

  Our work focuses on the development of a novel camera localization method that leverages high-level map primitives under a unified and scalable framework, which is highly suitable for memory-constrained and real-time applications such as AR/VR and robotics.
  However, with the rapid growth in the popularity of cloud-based localization services, potential privacy concerns arise during localization, as users’ scene-related information may be exposed to unauthorized access or misuse.
  One possible mitigation is to anonymize sensitive scene data on-device, for instance by tokenizing the query image and the map prior to transmission.


\begin{thebibliography}{100}
\providecommand{\natexlab}[1]{#1}
\providecommand{\url}[1]{\texttt{#1}}
\expandafter\ifx\csname urlstyle\endcsname\relax
  \providecommand{\doi}[1]{doi: #1}\else
  \providecommand{\doi}{doi: \begingroup \urlstyle{rm}\Url}\fi

\bibitem[Abonce et~al.(2020)Abonce, Zhou, and Calway]{abonceYouAreHere2020}
Obed No{\'e}~Samano Abonce, Mengjie Zhou, and Andrew Calway.
\newblock You are here: {{Geolocation}} by embedding maps and images.
\newblock In \emph{{{ECCV}}}, 2020.

\bibitem[ARCore(2026)]{arcoreFundamentalConceptsEnvironmental2026}
ARCore.
\newblock Fundamental {{Concepts}}: {{Environmental Understanding}}.
\newblock Accessed: 2026-04-25.
\newblock Available at
  \url{https://developers.google.com/ar/develop/fundamentals\#environmental_understanding}.

\bibitem[Armeni et~al.(2019)Armeni, He, Zamir, Gwak, Malik, Fischer, and
  Savarese]{armeni3DSceneGraph2019}
Iro Armeni, Zhi-Yang He, Amir Zamir, Junyoung Gwak, Jitendra Malik, Martin
  Fischer, and Silvio Savarese.
\newblock {{3D Scene Graph}}: {{A Structure}} for {{Unified Semantics}}, {{3D
  Space}}, and {{Camera}}.
\newblock In \emph{{{ICCV}}}. 2019.

\bibitem[Avetisyan et~al.(2020)Avetisyan, Khanova, Choy, Dash, Dai, and
  Nie{\ss}ner]{avetisyanSceneCADPredictingObject2020}
Armen Avetisyan, Tatiana Khanova, Christopher~Bongsoo Choy, Denver Dash, Angela
  Dai, and Matthias Nie{\ss}ner.
\newblock {{SceneCAD}}: {{Predicting}} object alignments and layouts in
  {{RGB}}-d scans.
\newblock In \emph{{{ECCV}}}, 2020.

\bibitem[Avetisyan et~al.(2024)Avetisyan, Xie, {Howard-Jenkins}, Yang, Aroudj,
  Patra, Zhang, Frost, Holland, Orme, Engel, Miller, Newcombe, and
  Balntas]{avetisyanSceneScriptReconstructingScenes2024}
Armen Avetisyan, Christopher Xie, Henry {Howard-Jenkins}, Tsun-Yi Yang, Samir
  Aroudj, Suvam Patra, Fuyang Zhang, Duncan Frost, Luke Holland, Campbell Orme,
  Jakob Engel, Edward Miller, Richard Newcombe, and Vasileios Balntas.
\newblock {{SceneScript}}: {{Reconstructing Scenes With An Autoregressive
  Structured Language Model}}.
\newblock In \emph{{{ECCV}}}. 2024.

\bibitem[{Barroso-Laguna} et~al.(2026){Barroso-Laguna}, Cavallari, Prisacariu,
  and Brachmann]{barroso-lagunaSceneWorthThousand2025}
Axel {Barroso-Laguna}, Tommaso Cavallari, Victor~Adrian Prisacariu, and Eric
  Brachmann.
\newblock A {{Scene}} is {{Worth}} a {{Thousand Features}}: {{Feed-Forward
  Camera Localization}} from a {{Collection}} of {{Image Features}}.
\newblock In \emph{{{ICLR}}}. 2026.

\bibitem[{Blender Online
  Community}(2026)]{blenderonlinecommunityBlender3DModelling}
{Blender Online Community}.
\newblock Blender - a {{3D}} modelling and rendering package.
\newblock 2026.
\newblock Available at \url{http://www.blender.org}.

\bibitem[Brachmann \& Rother(2019)Brachmann and
  Rother]{brachmannExpertSampleConsensus2019}
Eric Brachmann and Carsten Rother.
\newblock Expert {{Sample Consensus Applied}} to {{Camera Re-Localization}}.
\newblock In \emph{{{ICCV}}}. 2019.

\bibitem[Brachmann et~al.(2016)Brachmann, Krull, Nowozin, Shotton, Michel,
  Gumhold, and Rother]{brachmannDSACDifferentiableRANSAC2016}
Eric Brachmann, Alexander Krull, Sebastian Nowozin, Jamie Shotton, Frank
  Michel, Stefan Gumhold, and Carsten Rother.
\newblock {{DSAC}} --- differentiable {{RANSAC}} for camera localization.
\newblock In \emph{{{CVPR}}}, 2016.

\bibitem[Brachmann et~al.(2023)Brachmann, Cavallari, and
  Prisacariu]{brachmannAcceleratedCoordinateEncoding2023}
Eric Brachmann, Tommaso Cavallari, and Victor~Adrian Prisacariu.
\newblock Accelerated coordinate encoding: {{Learning}} to relocalize in
  minutes using {{RGB}} and poses.
\newblock In \emph{{{CVPR}}}. 2023.

\bibitem[Camposeco et~al.(2019)Camposeco, Cohen, Pollefeys, and
  Sattler]{camposecoHybridSceneCompression2019}
Federico Camposeco, Andrea Cohen, Marc Pollefeys, and Torsten Sattler.
\newblock Hybrid scene compression for visual localization.
\newblock In \emph{{{CVPR}}}. 2019.

\bibitem[Chang et~al.(2017)Chang, Dai, Funkhouser, Halber, Nie{\ss}ner, Savva,
  Song, Zeng, and Zhang]{changMatterport3DLearningRGBD2017}
Angel Chang, Angela Dai, Thomas Funkhouser, Maciej Halber, Matthias
  Nie{\ss}ner, Manolis Savva, Shuran Song, Andy Zeng, and Yinda Zhang.
\newblock {{Matterport3D}}: {{Learning}} from {{RGB-D Data}} in {{Indoor
  Environments}}.
\newblock In \emph{{{3DV}}}. 2017.

\bibitem[Charatan et~al.(2024)Charatan, Li, Tagliasacchi, and
  Sitzmann]{charatan23pixelsplat}
David Charatan, Sizhe Li, Andrea Tagliasacchi, and Vincent Sitzmann.
\newblock {{pixelSplat}}: {{3D}} gaussian splats from image pairs for scalable
  generalizable {{3D}} reconstruction.
\newblock In \emph{{{CVPR}}}, 2024.

\bibitem[Chen et~al.(2025)Chen, Kang, Yang, Zhong, and
  Wang]{chenPerspectiveHigherDimension2025}
Bolei Chen, Jiaxu Kang, Haonan Yang, Ping Zhong, and Jianxin Wang.
\newblock Perspective from a {{Higher Dimension}}: {{Can 3D Geometric Priors
  Help Visual Floorplan Localization}}?
\newblock In \emph{{{ACM MM}}}. 2025.

\bibitem[Chen et~al.(2026{\natexlab{a}})Chen, Yan, Cui, Kang, Zhong, and
  Wang]{chenPerspectiveBroaderContext2025}
Bolei Chen, Shengsheng Yan, Yongzheng Cui, Jiaxu Kang, Ping Zhong, and Jianxin
  Wang.
\newblock Perspective from a {{Broader Context}}: {{Can Room Style Knowledge
  Help Visual Floorplan Localization}}?
\newblock In \emph{{{AAAI}}}. 2026{\natexlab{a}}.

\bibitem[Chen et~al.(2024{\natexlab{a}})Chen, Wang, Vogel, and
  Pollefeys]{chenF$^3$LocFusionFiltering2024}
Changan Chen, Rui Wang, Christoph Vogel, and Marc Pollefeys.
\newblock F\$\textasciicircum 3\${{Loc}}: {{Fusion}} and {{Filtering}} for
  {{Floorplan Localization}}.
\newblock In \emph{{{CVPR}}}. 2024{\natexlab{a}}.

\bibitem[Chen et~al.(2024{\natexlab{b}})Chen, Bhalgat, Li, Bian, Li, Wang, and
  Prisacariu]{chenRefinementAbsolutePose2024}
Shuai Chen, Yash Bhalgat, Xing~Hui Li, Jia~Wang Bian, Ke~Jie Li, Zirui Wang,
  and Victor~Adrian Prisacariu.
\newblock Refinement for {{Absolute Pose Regression}} with {{Neural Feature
  Synthesis}}.
\newblock In \emph{{{CVPR}}}. 2024{\natexlab{b}}.

\bibitem[Chen et~al.(2026{\natexlab{b}})Chen, Chen, Xiu, Geiger, and
  Chen]{chenTTT3R3DReconstruction2026}
Xingyu Chen, Yue Chen, Yuliang Xiu, Andreas Geiger, and Anpei Chen.
\newblock {{TTT3R}}: {{3D Reconstruction}} as {{Test-Time Training}}.
\newblock In \emph{{{ICLR}}}. 2026{\natexlab{b}}.

\bibitem[Cruz et~al.(2021)Cruz, Hutchcroft, Li, Khosravan, Boyadzhiev, and
  Kang]{cruzZillowIndoorDataset2021}
Steve Cruz, Will Hutchcroft, Yuguang Li, Naji Khosravan, Ivaylo Boyadzhiev, and
  Sing~Bing Kang.
\newblock Zillow indoor dataset: {{Annotated}} floor plans with 360{$^\circ$}
  panoramas and {{3D}} room layouts.
\newblock In \emph{{{CVPR}}}, 2021.

\bibitem[Dai et~al.(2017)Dai, Chang, Savva, Halber, Funkhouser, and
  Nie{\ss}ner]{daiScanNetRichlyannotated3D2017}
Angela Dai, Angel~X. Chang, Manolis Savva, Maciej Halber, Thomas Funkhouser,
  and Matthias Nie{\ss}ner.
\newblock {{ScanNet}}: {{Richly}}-annotated {{3D}} reconstructions of indoor
  scenes.
\newblock In \emph{{{CVPR}}}, 2017.

\bibitem[Dong et~al.(2025)Dong, Wang, Liu, Cai, Fan, Kannala, and
  Yang]{dongReloc3rLargeScaleTraining2025}
Siyan Dong, Shuzhe Wang, Shaohui Liu, Lulu Cai, Qingnan Fan, Juho Kannala, and
  Yanchao Yang.
\newblock Reloc3r: {{Large-Scale Training}} of {{Relative Camera Pose
  Regression}} for {{Generalizable}}, {{Fast}}, and {{Accurate Visual
  Localization}}.
\newblock In \emph{{{CVPR}}}. 2025.

\bibitem[Dosovitskiy et~al.(2021)Dosovitskiy, Beyer, Kolesnikov, Weissenborn,
  Zhai, Unterthiner, Dehghani, Minderer, Heigold, Gelly, Uszkoreit, and
  Houlsby]{dosovitskiyImageWorth16x162021}
Alexey Dosovitskiy, Lucas Beyer, Alexander Kolesnikov, Dirk Weissenborn,
  Xiaohua Zhai, Thomas Unterthiner, Mostafa Dehghani, Matthias Minderer, Georg
  Heigold, Sylvain Gelly, Jakob Uszkoreit, and Neil Houlsby.
\newblock An {{Image}} is {{Worth}} 16x16 {{Words}}: {{Transformers}} for
  {{Image Recognition}} at {{Scale}}.
\newblock In \emph{{{ICLR}}}. 2021.

\bibitem[Fischler \& Bolles(1981)Fischler and
  Bolles]{fischlerRandomSampleConsensus1981}
Martin~A. Fischler and Robert~C. Bolles.
\newblock Random sample consensus: A paradigm for model fitting with
  applications to image analysis and automated cartography.
\newblock \emph{Communications of The Acm}, 24\penalty0 (6):\penalty0 381--395,
  1981.

\bibitem[Gao et~al.(2003)Gao, Hou, Tang, and
  Cheng]{gaoCompleteSolutionClassification2003}
Xiaoshan Gao, Xiaorong Hou, Jianliang Tang, and Hangfei Cheng.
\newblock Complete solution classification for the perspective-three-point
  problem.
\newblock \emph{IEEE Trans. Pattern Anal. Mach. Intell.}, 25\penalty0
  (8):\penalty0 930--943, 2003.

\bibitem[Gard et~al.(2024)Gard, Hilsmann, and
  Eisert]{gardSPVLocSemanticPanoramic2024}
Niklas Gard, Anna Hilsmann, and Peter Eisert.
\newblock {{SPVLoc}}: {{Semantic Panoramic Viewport Matching}} for {{6D Camera
  Localization}} in {{Unseen Environments}}.
\newblock In \emph{{{ECCV}}}. 2024.

\bibitem[{GEOS
  contributors}(2025)]{geoscontributorsGEOSComputationalGeometry2025}
{GEOS contributors}.
\newblock {{GEOS}} computational geometry library.
\newblock 2025.
\newblock Available at \url{https://libgeos.org/}.

\bibitem[Grader \& {Averbuch-Elor}(2025)Grader and
  {Averbuch-Elor}]{graderSuperchargingFloorplanLocalization2025}
Yuval Grader and Hadar {Averbuch-Elor}.
\newblock Supercharging floorplan localization with semantic rays.
\newblock In \emph{{{ICCV}}}, 2025.

\bibitem[Guo et~al.(2021)Guo, Lin, Guo, Liang, and
  Cheng]{Guo2021CoarsetofineSL}
Chengcheng Guo, Minjie Lin, Heyang Guo, Pengpeng Liang, and Erkang Cheng.
\newblock Coarse-to-fine semantic localization with {{HD}} map for autonomous
  driving in structural scenes.
\newblock In \emph{{{IROS}}}, 2021.

\bibitem[Hanlon et~al.(2024)Hanlon, Sun, Pollefeys, and
  Blum]{hanlonActiveVisualLocalization2024}
Matthew Hanlon, Boyang Sun, Marc Pollefeys, and Hermann Blum.
\newblock Active {{Visual Localization}} for {{Multi-Agent Collaboration}}: {{A
  Data-Driven Approach}}.
\newblock In \emph{{{ICRA}}}. 2024.

\bibitem[He et~al.(2025)He, Yu, Peng, Tan, Shen, Bao, and
  Zhou]{heMatchAnythingUniversalCrossModality2025}
Xingyi He, Hao Yu, Sida Peng, Dongli Tan, Zehong Shen, Hujun Bao, and Xiaowei
  Zhou.
\newblock {{MatchAnything}}: {{Universal Cross-Modality Image Matching}} with
  {{Large-Scale Pre-Training}}, \emph{arXiv preprint arXiv:}\emph{2501.07556},
  2025.
\newblock Available at \url{http://arxiv.org/abs/2501.07556}.

\bibitem[Hong et~al.(2019)Hong, P{\'e}tillot, Lane, Miao, and
  Wang]{Hong2019TextPlaceVP}
Ziyang Hong, Yvan~R. P{\'e}tillot, David~M. Lane, Yishu Miao, and Sen Wang.
\newblock {{TextPlace}}: {{Visual}} place recognition and topological
  localization through reading scene texts.
\newblock \emph{ICCV}, pp.\  2861--2870, 2019.

\bibitem[{Howard-Jenkins} et~al.(2021){Howard-Jenkins}, {Ruiz-Sarmiento}, and
  Prisacariu]{howard-jenkinsLalalocLatentLayout2021}
Henry {Howard-Jenkins}, Jose-Raul {Ruiz-Sarmiento}, and Victor~Adrian
  Prisacariu.
\newblock Lalaloc: {{Latent}} layout localisation in dynamic, unvisited
  environments.
\newblock In \emph{{{ICCV}}}, 2021.

\bibitem[Huang et~al.(2025)Huang, Yu, Shentu, Yuan, and
  Zhang]{huangSparseDenseCamera2025}
Zhiwei Huang, Hailin Yu, Yichun Shentu, Jin Yuan, and Guofeng Zhang.
\newblock From sparse to dense: {{Camera}} relocalization with scene-specific
  detector from {{Feature Gaussian Splatting}}.
\newblock In \emph{{{CVPR}}}, 2025.

\bibitem[Humenberger et~al.(2022)Humenberger, Cabon, Guerin, Morat, Leroy,
  Revaud, Rerole, Pion, de~Souza, and
  Csurka]{humenbergerRobustImageRetrievalbased2022}
Martin Humenberger, Yohann Cabon, Nicolas Guerin, Julien Morat, Vincent Leroy,
  J{\'e}r{\^o}me Revaud, Philippe Rerole, No{\'e} Pion, Cesar de~Souza, and
  Gabriela Csurka.
\newblock Robust {{Image Retrieval}}-based {{Visual Localization}} using
  {{Kapture}}, \emph{arXiv preprint arXiv:}\emph{2007.13867}, 2022.
\newblock Available at \url{http://arxiv.org/abs/2007.13867}.

\bibitem[Jang et~al.(2025)Jang, Weinzaepfel, Leroy, Agapito, and
  Revaud]{jangPow3REmpoweringUnconstrained2025}
Wonbong Jang, Philippe Weinzaepfel, Vincent Leroy, Lourdes Agapito, and Jerome
  Revaud.
\newblock {{Pow3R}}: {{Empowering}} unconstrained {{3D}} reconstruction with
  camera and scene priors.
\newblock In \emph{{{CVPR}}}, 2025.

\bibitem[Jiang et~al.(2025)Jiang, Wang, Galliani, Vogel, and
  Pollefeys]{jiangRSCoReRevisitingScene2025}
Xudong Jiang, Fangjinhua Wang, Silvano Galliani, Christoph Vogel, and Marc
  Pollefeys.
\newblock R-{{SCoRe}}: {{Revisiting Scene Coordinate Regression}} for {{Robust
  Large-Scale Visual Localization}}.
\newblock In \emph{{{CVPR}}}. 2025.

\bibitem[Jin et~al.(2021)Jin, Qian, Owens, and
  Fouhey]{jinPlanarSurfaceReconstruction2021}
Linyi Jin, Shengyi Qian, Andrew Owens, and David~F. Fouhey.
\newblock Planar surface reconstruction from sparse views.
\newblock In \emph{{{ICCV}}}. 2021.

\bibitem[Karkus et~al.(2018)Karkus, Hsu, and
  Lee]{karkusParticleFilterNetworks2018}
Peter Karkus, David Hsu, and Wee~Sun Lee.
\newblock Particle filter networks with application to visual localization.
\newblock In \emph{{{CoRL}}}. 2018.

\bibitem[Keetha et~al.(2025)Keetha, M{\"u}ller, Sch{\"o}nberger, Porzi, Zhang,
  Fischer, Knapitsch, Zauss, Weber, Antunes, Luiten, {Lopez-Antequera},
  Bul{\`o}, Richardt, Ramanan, Scherer, and
  Kontschieder]{keethaMapAnythingUniversalFeedForward2025}
Nikhil Keetha, Norman M{\"u}ller, Johannes Sch{\"o}nberger, Lorenzo Porzi,
  Yuchen Zhang, Tobias Fischer, Arno Knapitsch, Duncan Zauss, Ethan Weber,
  Nelson Antunes, Jonathon Luiten, Manuel {Lopez-Antequera}, Samuel~Rota
  Bul{\`o}, Christian Richardt, Deva Ramanan, Sebastian Scherer, and Peter
  Kontschieder.
\newblock {{MapAnything}}: {{Universal Feed-Forward Metric 3D Reconstruction}}.
\newblock In \emph{{{3DV}}}. 2025.

\bibitem[Larsson \& {contributors}(2020)Larsson and
  {contributors}]{larssonPoseLibMinimalSolvers2020}
Viktor Larsson and {contributors}.
\newblock {{PoseLib}} - minimal solvers for camera pose estimation.
\newblock 2020.
\newblock Available at \url{https://github.com/vlarsson/PoseLib}.

\bibitem[Leroy et~al.(2024)Leroy, Cabon, and
  Revaud]{leroyGroundingImageMatching2024}
Vincent Leroy, Yohann Cabon, and J{\'e}r{\^o}me Revaud.
\newblock Grounding {{Image Matching}} in {{3D}} with {{MASt3R}}.
\newblock In \emph{{{ECCV}}}. 2024.

\bibitem[Li et~al.(2025)Li, Sun, Giammarino, Blum, and
  Pollefeys]{liActLocLearningLocalize2025}
Jiajie Li, Boyang Sun, Luca~Di Giammarino, Hermann Blum, and Marc Pollefeys.
\newblock {{ActLoc}}: {{Learning}} to {{Localize}} on the {{Move}} via {{Active
  Viewpoint Selection}}.
\newblock In \emph{{{CoRL}}}. 2025.

\bibitem[Lin et~al.(2026)Lin, Chen, Liew, Chen, Li, Shi, Feng, and
  Kang]{linDepthAnything32026}
Haotong Lin, Sili Chen, Junhao Liew, Donny~Y. Chen, Zhenyu Li, Guang Shi,
  Jiashi Feng, and Bingyi Kang.
\newblock Depth {{Anything}} 3: {{Recovering}} the {{Visual Space}} from {{Any
  Views}}.
\newblock In \emph{{{ICLR}}}, 2026.

\bibitem[Liu et~al.(2025{\natexlab{a}})Liu, Tan, Ke, Zhang, Liu, Qian, Xue,
  Shen, and Braud]{liuPLANA3RZeroshotMetric2025}
Changkun Liu, Bin Tan, Zeran Ke, Shangzhan Zhang, Jiachen Liu, Ming Qian, Nan
  Xue, Yujun Shen, and Tristan Braud.
\newblock {{PLANA3R}}: {{Zero}}-shot {{Metric Planar 3D Reconstruction}} via
  {{Feed-Forward Planar Splatting}}.
\newblock In \emph{{{NeurIPS}}}. 2025{\natexlab{a}}.

\bibitem[Liu et~al.(2018)Liu, Yang, Ceylan, Yumer, and
  Furukawa]{liuPlaneNetPiecewisePlanar2018}
Chen Liu, Jimei Yang, Duygu Ceylan, Ersin Yumer, and Yasutaka Furukawa.
\newblock {{PlaneNet}}: {{Piece}}-wise planar reconstruction from a single
  {{RGB}} image.
\newblock In \emph{{{CVPR}}}, 2018.

\bibitem[Liu et~al.(2019)Liu, Kim, Gu, Furukawa, and
  Kautz]{liuPlaneRCNN3DPlane2019}
Chen Liu, Kihwan Kim, Jinwei Gu, Yasutaka Furukawa, and Jan Kautz.
\newblock {{PlaneRCNN}}: {{3D}} plane detection and reconstruction from a
  single image.
\newblock In \emph{{{CVPR}}}. 2019.

\bibitem[Liu et~al.(2025{\natexlab{b}})Liu, Yu, Chen, Huang, and
  Guo]{liuInthewild3DPlane2025}
Jiachen Liu, Rui Yu, Sili Chen, Sharon~X. Huang, and Hengkai Guo.
\newblock Towards {{In}}-the-wild {{3D Plane Reconstruction}} from a {{Single
  Image}}.
\newblock In \emph{{{CVPR}}}. 2025{\natexlab{b}}.

\bibitem[Liu et~al.(2017)Liu, Li, and Dai]{liuEfficientGlobal2d3d2017}
Liu Liu, Hongdong Li, and Yuchao Dai.
\newblock Efficient global 2d-3d matching for camera localization in a
  large-scale 3d map.
\newblock In \emph{{{ICCV}}}, 2017.

\bibitem[Liu et~al.(2024)Liu, Zhu, Ma, Ye, Gao, Zheng, and
  Shen]{liuPolyRoomRoomawareTransformer2024}
Yuzhou Liu, Lingjie Zhu, Xiaodong Ma, Hanqiao Ye, Xiang Gao, Xianwei Zheng, and
  Shuhan Shen.
\newblock {{PolyRoom}}: {{Room}}-aware transformer for floorplan
  reconstruction.
\newblock In \emph{{{ECCV}}}. 2024.

\bibitem[Liu et~al.(2026)Liu, Zhu, Ye, Liu, Huang, Gao, Wang, and
  Shen]{liuBuildingGPTAutoRegressiveBuilding2026}
Yuzhou Liu, Lingjie Zhu, Hanqiao Ye, Yujun Liu, Shangfeng Huang, Xiang Gao,
  Ruisheng Wang, and Shuhan Shen.
\newblock {{BuildingGPT}}: {{Auto-Regressive Building Wireframe Reconstruction
  Model}} with {{Reinforcement Learning}}.
\newblock In \emph{{{CVPR}}}, 2026.

\bibitem[Loshchilov \& Hutter(2019)Loshchilov and
  Hutter]{loshchilovDecoupledWeightDecay2019}
Ilya Loshchilov and Frank Hutter.
\newblock Decoupled weight decay regularization.
\newblock In \emph{{{ICLR}}}, 2019.

\bibitem[Min et~al.(2022)Min, Khosravan, Bessinger, Narayana, Kang, Dunn, and
  Boyadzhiev]{minLASERLAtentSpacE2022}
Zhixiang Min, Naji Khosravan, Zachary Bessinger, Manjunath Narayana, Sing~Bing
  Kang, Enrique Dunn, and Ivaylo Boyadzhiev.
\newblock {{LASER}}: {{LAtent SpacE Rendering}} for {{2D Visual Localization}}.
\newblock In \emph{{{CVPR}}}. 2022.

\bibitem[Monszpart et~al.(2015)Monszpart, Mellado, Brostow, and
  Mitra]{monszpartRAPterRebuildingManmade2015}
{\'A}ron Monszpart, Nicolas Mellado, Gabriel~J. Brostow, and Niloy~J. Mitra.
\newblock {{RAPter}}: Rebuilding man-made scenes with regular arrangements of
  planes.
\newblock \emph{ACM Trans. Graph.}, 34\penalty0 (4):\penalty0 103:1--103:12,
  2015.

\bibitem[Moreau et~al.(2023)Moreau, Piasco, Bennehar, Tsishkou, Stanciulescu,
  and {de La Fortelle}]{moreauCROSSFIRECameraRelocalization2023}
Arthur Moreau, Nathan Piasco, Moussab Bennehar, Dzmitry Tsishkou, Bogdan
  Stanciulescu, and Arnaud {de La Fortelle}.
\newblock {{CROSSFIRE}}: {{Camera}} relocalization on self-supervised features
  from an implicit representation.
\newblock In \emph{{{ICCV}}}. 2023.

\bibitem[Murai et~al.(2025)Murai, Dexheimer, and
  Davison]{muraiMASt3RSLAMRealTimeDense2025}
Riku Murai, Eric Dexheimer, and Andrew~J. Davison.
\newblock {{MASt3R-SLAM}}: {{Real-Time Dense SLAM}} with {{3D Reconstruction
  Priors}}.
\newblock In \emph{{{CVPR}}}. 2025.

\bibitem[Oquab et~al.(2024)Oquab, Darcet, Moutakanni, Vo, Szafraniec, Khalidov,
  Fernandez, Haziza, Massa, {El-Nouby}, Assran, Ballas, Galuba, Howes, Huang,
  Li, Misra, Rabbat, Sharma, Synnaeve, Xu, Jegou, Mairal, Labatut, Joulin, and
  Bojanowski]{oquabDINOv2LearningRobust2023}
Maxime Oquab, Timoth{\'e}e Darcet, Th{\'e}o Moutakanni, Huy Vo, Marc
  Szafraniec, Vasil Khalidov, Pierre Fernandez, Daniel Haziza, Francisco Massa,
  Alaaeldin {El-Nouby}, Mahmoud Assran, Nicolas Ballas, Wojciech Galuba,
  Russell Howes, Po-Yao Huang, Shang-Wen Li, Ishan Misra, Michael Rabbat, Vasu
  Sharma, Gabriel Synnaeve, Hu~Xu, Herv{\'e} Jegou, Julien Mairal, Patrick
  Labatut, Armand Joulin, and Piotr Bojanowski.
\newblock {{DINOv2}}: {{Learning Robust Visual Features}} without
  {{Supervision}}.
\newblock \emph{TMLR}, pp.\  2835--8856, 2024.

\bibitem[Panek et~al.(2022)Panek, Kukelova, and
  Sattler]{panekMeshLocMeshbasedVisual2022}
Vojtech Panek, Zuzana Kukelova, and Torsten Sattler.
\newblock {{MeshLoc}}: {{Mesh}}-based visual localization.
\newblock In \emph{{{ECCV}}}, 2022.

\bibitem[Panek et~al.(2023)Panek, Kukelova, and
  Sattler]{panekVisualLocalizationUsing2023}
Vojtech Panek, Zuzana Kukelova, and Torsten Sattler.
\newblock Visual {{Localization}} using {{Imperfect 3D Models}} from the
  {{Internet}}.
\newblock In \emph{{{ICCV}}}. 2023.

\bibitem[Panphattarasap \& Calway(2018)Panphattarasap and
  Calway]{Panphattarasap2018AutomatedMR}
Pilailuck Panphattarasap and Andrew Calway.
\newblock Automated map reading: {{Image}} based localisation in 2-{{D}} maps
  using binary semantic descriptors.
\newblock In \emph{{{IROS}}}, 2018.

\bibitem[Qi et~al.(2017)Qi, Su, Mo, and Guibas]{qiPointNetDeepLearning2017}
Charles~R. Qi, Hao Su, Kaichun Mo, and Leonidas~J. Guibas.
\newblock {{PointNet}}: {{Deep Learning}} on {{Point Sets}} for {{3D
  Classification}} and {{Segmentation}}.
\newblock In \emph{{{CVPR}}}. 2017.

\bibitem[Qin et~al.(2023)Qin, Yu, Wang, Guo, Peng, Ilic, Hu, and
  Xu]{qinGeoTransformerFastRobust2023}
Zheng Qin, Hao Yu, Changjian Wang, Yulan Guo, Yuxing Peng, Slobodan Ilic, Dewen
  Hu, and Kai Xu.
\newblock {{GeoTransformer}}: {{Fast}} and {{Robust Point Cloud Registration}}
  with {{Geometric Transformer}}.
\newblock \emph{IEEE Trans. Pattern Anal. Mach. Intell.}, 45\penalty0
  (8):\penalty0 9806--9821, 2023.

\bibitem[Sarlin et~al.(2019)Sarlin, Cadena, Siegwart, and
  Dymczyk]{sarlinCoarseFineRobust2019}
Paul-Edouard Sarlin, Cesar Cadena, Roland Siegwart, and Marcin Dymczyk.
\newblock From {{Coarse}} to {{Fine}}: {{Robust Hierarchical Localization}} at
  {{Large Scale}}.
\newblock In \emph{{{CVPR}}}. 2019.

\bibitem[Sarlin et~al.(2023{\natexlab{a}})Sarlin, DeTone, Yang, Avetisyan,
  Straub, Malisiewicz, Bul{\`o}, Newcombe, Kontschieder, and
  Balntas]{sarlinOrienterNetVisualLocalization2023}
Paul-Edouard Sarlin, Daniel DeTone, Tsun-Yi Yang, Armen Avetisyan, Julian
  Straub, Tomasz Malisiewicz, Samuel~Rota Bul{\`o}, Richard Newcombe, Peter
  Kontschieder, and Vasileios Balntas.
\newblock {{OrienterNet}}: {{Visual}} localization in {{2D}} public maps with
  neural matching.
\newblock In \emph{{{CVPR}}}, 2023{\natexlab{a}}.

\bibitem[Sarlin et~al.(2023{\natexlab{b}})Sarlin, Trulls, Pollefeys, Hosang,
  and Lynen]{sarlinSNAPSelfsupervisedNeural2023}
Paul-Edouard Sarlin, Eduard Trulls, Marc Pollefeys, Jan Hosang, and Simon
  Lynen.
\newblock {{SNAP}}: {{Self}}-supervised neural maps for visual positioning and
  semantic understanding.
\newblock In \emph{{{NeurIPS}}}, 2023{\natexlab{b}}.

\bibitem[Savva et~al.(2019)Savva, Kadian, Maksymets, Zhao, Wijmans, Jain,
  Straub, Liu, Koltun, Malik, Parikh, and
  Batra]{savvaHabitatPlatformEmbodied2019}
Manolis Savva, Abhishek Kadian, Oleksandr Maksymets, Yili Zhao, Erik Wijmans,
  Bhavana Jain, Julian Straub, Jia Liu, Vladlen Koltun, Jitendra Malik, Devi
  Parikh, and Dhruv Batra.
\newblock Habitat: {{A Platform}} for {{Embodied AI Research}}.
\newblock In \emph{{{ICCV}}}, 2019.

\bibitem[Sch{\"o}nberger et~al.(2018)Sch{\"o}nberger, Pollefeys, Geiger, and
  Sattler]{schonbergerSemanticVisualLocalization2018}
Johannes~L. Sch{\"o}nberger, Marc Pollefeys, Andreas Geiger, and Torsten
  Sattler.
\newblock Semantic {{Visual Localization}}.
\newblock In \emph{{{CVPR}}}. 2018.

\bibitem[Shi et~al.(2023)Shi, Zhi, and Xu]{shiPlaneRecTRUnifiedQuery2023}
Jingjia Shi, Shuaifeng Zhi, and Kai Xu.
\newblock {{PlaneRecTR}}: {{Unified Query Learning}} for {{3D Plane Recovery}}
  from a {{Single View}}.
\newblock In \emph{{{ICCV}}}, 2023.

\bibitem[Shi et~al.(2026)Shi, Zhi, and Xu]{shiPlaneRecTRUnifiedQuery2026}
Jingjia Shi, Shuaifeng Zhi, and Kai Xu.
\newblock {{PlaneRecTR}}++: {{Unified Query Learning}} for {{Joint 3D Planar
  Reconstruction}} and {{Pose Estimation}}.
\newblock \emph{IEEE Trans. Pattern Anal. Mach. Intell.}, 48\penalty0
  (1):\penalty0 962--981, 2026.

\bibitem[Shotton et~al.(2013)Shotton, Glocker, Zach, Izadi, Criminisi, and
  Fitzgibbon]{shottonSceneCoordinateRegression2013}
Jamie Shotton, Ben Glocker, Christopher Zach, Shahram Izadi, Antonio Criminisi,
  and Andrew Fitzgibbon.
\newblock Scene coordinate regression forests for camera relocalization in
  {{RGB}}-d images.
\newblock In \emph{{{CVPR}}}, 2013.

\bibitem[Smart et~al.(2024)Smart, Zheng, Laina, and
  Prisacariu]{smart2024splatt3r}
Brandon Smart, Chuanxia Zheng, Iro Laina, and Victor~Adrian Prisacariu.
\newblock {{Splatt3R}}: {{Zero}}-shot gaussian splatting from uncalibrated
  image pairs, \emph{arXiv preprint arXiv:}\emph{2408.13912}, 2024.
\newblock Available at \url{https://arxiv.org/abs/2408.13912}.

\bibitem[Stary et~al.(2025)Stary, Gaubil, Tewari, and
  Sitzmann]{staryUnderstandingMultiViewTransformers2025}
Michal Stary, Julien Gaubil, Ayush Tewari, and Vincent Sitzmann.
\newblock Understanding {{Multi-View Transformers}}.
\newblock In \emph{{{ICCVW}}}. 2025.

\bibitem[Taira et~al.(2018)Taira, Okutomi, Sattler, Cimpoi, Pollefeys, Sivic,
  Pajdla, and Torii]{tairaInLocIndoorVisual2018}
Hajime Taira, Masatoshi Okutomi, Torsten Sattler, Mircea Cimpoi, Marc
  Pollefeys, Josef Sivic, Tomas Pajdla, and Akihiko Torii.
\newblock {{InLoc}}: {{Indoor Visual Localization}} with {{Dense Matching}} and
  {{View Synthesis}}.
\newblock In \emph{{{CVPR}}}. 2018.

\bibitem[Tan et~al.(2023)Tan, Xue, Wu, and Xia]{tanNOPESACNeuralOneplane2023}
Bin Tan, Nan Xue, Tianfu Wu, and Gui-Song Xia.
\newblock {{NOPE-SAC}}: {{Neural}} one-plane {{RANSAC}} for sparse-view planar
  {{3D}} reconstruction.
\newblock \emph{IEEE Trans. Pattern Anal. Mach. Intell.}, 45\penalty0
  (12):\penalty0 15233--15248, 2023.

\bibitem[Tan et~al.(2025)Tan, Yu, Shen, and
  Xue]{tanPlanarSplattingAccuratePlanar2025}
Bin Tan, Rui Yu, Yujun Shen, and Nan Xue.
\newblock {{PlanarSplatting}}: {{Accurate Planar Surface Reconstruction}} in 3
  {{Minutes}}.
\newblock In \emph{{{CVPR}}}. 2025.

\bibitem[Valentin et~al.(2016)Valentin, Dai, Nie{\ss}ner, Kohli, Torr, Izadi,
  and Keskin]{valentinLearningNavigateEnergy2016}
Julien Valentin, Angela Dai, Matthias Nie{\ss}ner, Pushmeet Kohli, Philip Torr,
  Shahram Izadi, and Cem Keskin.
\newblock Learning to navigate the energy landscape.
\newblock In \emph{{{3DV}}}. 2016.

\bibitem[Vojir et~al.(2020)Vojir, Budvytis, and
  Cipolla]{vojirEfficientLargescaleSemantic2020}
Tomas Vojir, Ignas Budvytis, and Roberto Cipolla.
\newblock Efficient large-scale semantic visual localization in {{2D}} maps.
\newblock In \emph{{{ACCV}}}, 2020.

\bibitem[Wang et~al.(2024{\natexlab{a}})Wang, Jiang, Galliani, Vogel, and
  Pollefeys]{wangGLACEGlobalLocal2024}
Fangjinhua Wang, Xudong Jiang, Silvano Galliani, Christoph Vogel, and Marc
  Pollefeys.
\newblock {{GLACE}}: {{Global}} local accelerated coordinate encoding.
\newblock In \emph{{{CVPR}}}, 2024{\natexlab{a}}.

\bibitem[Wang et~al.(2024{\natexlab{b}})Wang, Liu, Wang, Sun, Dong, Wang, and
  Yang]{wangFreeRegImagetoPointCloud2024}
Haiping Wang, Yuan Liu, Bing Wang, Yujing Sun, Zhen Dong, Wenping Wang, and
  Bisheng Yang.
\newblock {{FreeReg}}: {{Image}}-to-{{Point Cloud Registration Leveraging
  Pretrained Diffusion Models}} and {{Monocular Depth Estimators}}.
\newblock In \emph{{{ICLR}}}, 2024{\natexlab{b}}.

\bibitem[Wang et~al.(2025{\natexlab{a}})Wang, Chen, Karaev, Vedaldi, Rupprecht,
  and Novotny]{wangVGGTVisualGeometry2025}
Jianyuan Wang, Minghao Chen, Nikita Karaev, Andrea Vedaldi, Christian
  Rupprecht, and David Novotny.
\newblock {{VGGT}}: {{Visual Geometry Grounded Transformer}}.
\newblock In \emph{{{CVPR}}}. 2025{\natexlab{a}}.

\bibitem[Wang et~al.(2025{\natexlab{b}})Wang, Zhang, Holynski, Efros, and
  Kanazawa]{wangContinuous3DPerception2025}
Qianqian Wang, Yifei Zhang, Aleksander Holynski, Alexei~A. Efros, and Angjoo
  Kanazawa.
\newblock Continuous {{3D Perception Model}} with {{Persistent State}}.
\newblock In \emph{{{CVPR}}}. 2025{\natexlab{b}}.

\bibitem[Wang et~al.(2025{\natexlab{c}})Wang, Xu, Dong, Deng, Xiang, Lv, Sun,
  Tong, and Yang]{wangMoGe2AccurateMonocular2025}
Ruicheng Wang, Sicheng Xu, Yue Dong, Yu~Deng, Jianfeng Xiang, Zelong Lv,
  Guangzhong Sun, Xin Tong, and Jiaolong Yang.
\newblock {{MoGe-2}}: {{Accurate Monocular Geometry}} with {{Metric Scale}} and
  {{Sharp Details}}.
\newblock In \emph{{{NeurIPS}}}. 2025{\natexlab{c}}.

\bibitem[Wang et~al.(2024{\natexlab{c}})Wang, Leroy, Cabon, Chidlovskii, and
  Revaud]{wangDUSt3RGeometric3D2024}
Shuzhe Wang, Vincent Leroy, Yohann Cabon, Boris Chidlovskii, and Jerome Revaud.
\newblock {{DUSt3R}}: {{Geometric 3D Vision Made Easy}}.
\newblock In \emph{{{CVPR}}}. 2024{\natexlab{c}}.

\bibitem[Watson et~al.(2024)Watson, Aleotti, Sayed, Qureshi, Mac~Aodha,
  Brostow, Firman, and Vicente]{watsonAirPlanesAccuratePlane2024}
Jamie Watson, Filippo Aleotti, Mohamed Sayed, Zawar Qureshi, Oisin Mac~Aodha,
  Gabriel Brostow, Michael Firman, and Sara Vicente.
\newblock {{AirPlanes}}: {{Accurate}} plane estimation via {{3D}}-consistent
  embeddings.
\newblock In \emph{{{CVPR}}}, 2024.

\bibitem[Wu et~al.(2024)Wu, Zhang, Lin, Mu, Zhao, Yang, and
  Qin]{wuMapLocNetCoarsetofineFeature2024}
Hang Wu, Zhenghao Zhang, Siyuan Lin, Xiangru Mu, Qiang Zhao, Ming Yang, and
  Tong Qin.
\newblock {{MapLocNet}}: {{Coarse}}-to-fine feature registration for visual
  re-localization in navigation maps.
\newblock In \emph{{{IROS}}}, 2024.

\bibitem[W{\"u}est et~al.(2026)W{\"u}est, Engelmann, Miksik, Pollefeys, and
  Barath]{wuestUnLocLeveragingDepth2026}
Matthias W{\"u}est, Francis Engelmann, Ondrej Miksik, Marc Pollefeys, and
  Daniel Barath.
\newblock {{UnLoc}}: {{Leveraging Depth Uncertainties}} for {{Floorplan
  Localization}}.
\newblock In \emph{{{ICLR}}}. 2026.

\bibitem[Xie et~al.(2022)Xie, Gadelha, Yang, Zhou, and
  Jiang]{xiePlanarReconRealtime3D2022}
Yiming Xie, Matheus Gadelha, Fengting Yang, Xiaowei Zhou, and Huaizu Jiang.
\newblock {{PlanarRecon}}: {{Realtime 3D Plane Detection}} and
  {{Reconstruction}} from {{Posed Monocular Videos}}.
\newblock In \emph{{{CVPR}}}, 2022.

\bibitem[Yang \& Zhou(2018)Yang and Zhou]{yangRecovering3DPlanes2018}
Fengting Yang and Zihan Zhou.
\newblock Recovering {{3D}} planes from a single image via convolutional neural
  networks.
\newblock In \emph{{{ECCV}}}, 2018.

\bibitem[Yang et~al.(2024)Yang, Jia, Zhi, and Huang]{yang2024physcene}
Yandan Yang, Baoxiong Jia, Peiyuan Zhi, and Siyuan Huang.
\newblock {{PhyScene}}: {{Physically}} interactable {{3D}} scene synthesis for
  embodied {{AI}}.
\newblock In \emph{{{CVPR}}}, 2024.

\bibitem[Ye et~al.(2025)Ye, Liu, Liu, and Shen]{ye2025neuralplane}
Hanqiao Ye, Yuzhou Liu, Yangdong Liu, and Shuhan Shen.
\newblock {{NeuralPlane}}: {{Structured 3D}} reconstruction in planar
  primitives with neural fields.
\newblock In \emph{{{ICLR}}}, 2025.

\bibitem[Ye et~al.(2026)Ye, Liu, Liu, and
  Shen]{yePlanaReLocCameraRelocalization2026}
Hanqiao Ye, Yuzhou Liu, Yangdong Liu, and Shuhan Shen.
\newblock {{PlanaReLoc}}: {{Camera Relocalization}} in {{3D Planar Primitives}}
  via {{Region-Based Structure Matching}}.
\newblock In \emph{{{CVPR}}}. 2026.

\bibitem[{Yen-Chen} et~al.(2021){Yen-Chen}, Florence, Barron, Rodriguez, Isola,
  and Lin]{yen-chenINeRFInvertingNeural2021}
Lin {Yen-Chen}, Pete Florence, Jonathan~T. Barron, Alberto Rodriguez, Phillip
  Isola, and Tsung-Yi Lin.
\newblock {{iNeRF}}: {{Inverting Neural Radiance Fields}} for {{Pose
  Estimation}}.
\newblock In \emph{2021 {{IEEE}}/{{RSJ International Conference}} on
  {{Intelligent Robots}} and {{Systems}}}. 2021.

\bibitem[Yeshwanth et~al.(2023)Yeshwanth, Liu, Nie{\ss}ner, and
  Dai]{yeshwanthScanNetHighfidelityDataset2023}
Chandan Yeshwanth, Yueh-Cheng Liu, Matthias Nie{\ss}ner, and Angela Dai.
\newblock {{ScanNet}}++: A high-fidelity dataset of {{3D}} indoor scenes.
\newblock In \emph{{{ICCV}}}, 2023.

\bibitem[Yu \& Lafarge(2022)Yu and Lafarge]{yuFindingGoodConfigurations2022}
Mulin Yu and Florent Lafarge.
\newblock Finding good configurations of planar primitives in unorganized point
  clouds.
\newblock In \emph{{{CVPR}}}. 2022.

\bibitem[Yu et~al.(2019)Yu, Zheng, Lian, Zhou, and
  Gao]{yuSingleimagePiecewisePlanar2019}
Zehao Yu, Jia Zheng, Dongze Lian, Zihan Zhou, and Shenghua Gao.
\newblock Single-image piece-wise planar {{3D}} reconstruction via associative
  embedding.
\newblock In \emph{{{CVPR}}}. 2019.

\bibitem[Zhai et~al.(2024)Zhai, Zhang, Zhao, Li, He, Cui, Bao, and
  Zhang]{zhaiSplatLoc3DGaussian2024}
Hongjia Zhai, Xiyu Zhang, Boming Zhao, Hai Li, Yijia He, Zhaopeng Cui, Hujun
  Bao, and Guofeng Zhang.
\newblock {{SplatLoc}}: {{3D Gaussian Splatting}}-based {{Visual Localization}}
  for {{Augmented Reality}}.
\newblock \emph{IEEE Trans. Vis. Comput. Graph.}, 2024.

\bibitem[Zhang et~al.(2025)Zhang, Herrmann, Hur, Jampani, Darrell, Cole, Sun,
  and Yang]{zhangMonST3RSimpleApproach2025}
Junyi Zhang, Charles Herrmann, Junhwa Hur, Varun Jampani, Trevor Darrell,
  Forrester Cole, Deqing Sun, and Ming-Hsuan Yang.
\newblock {{MonST3R}}: {{A Simple Approach}} for {{Estimating Geometry}} in the
  {{Presence}} of {{Motion}}.
\newblock In \emph{{{ICLR}}}. 2025.

\bibitem[Zheng et~al.(2020)Zheng, Zhang, Li, Tang, Gao, and
  Zhou]{zhengStructured3DLargePhotorealistic2020}
Jia Zheng, Junfei Zhang, Jing Li, Rui Tang, Shenghua Gao, and Zihan Zhou.
\newblock {{Structured3D}}: {{A Large Photo}}-realistic {{Dataset}} for
  {{Structured 3D Modeling}}.
\newblock In \emph{{{ECCV}}}. 2020.

\bibitem[Zhong et~al.(2026)Zhong, Meng, Zou, Chen, Mu, and
  Wang]{meng2026discofloc}
Ping Zhong, Shiyong Meng, Tao Zou, Bolei Chen, Chaoxu Mu, and Jianxin Wang.
\newblock {{DisCo}}-floc: {{Using}} dual-level visual-geometric contrasts to
  disambiguate depth-aware visual floorplan localization, 2026.
\newblock Available at \url{https://arxiv.org/abs/2601.01822}.

\bibitem[Zhu et~al.(2024)Zhu, Yan, Wang, {shengYue}, Liu, and
  Zhang]{zhuLoDlocAerialVisual2024}
Juelin Zhu, Shen Yan, Long Wang, zhang {shengYue}, Yu~Liu, and Maojun Zhang.
\newblock {{LoD}}-loc: {{Aerial}} visual localization using {{LoD 3D}} map with
  neural wireframe alignment.
\newblock In \emph{{{NeurIPS}}}. 2024.

\bibitem[Zhuo et~al.(2026)Zhuo, Zheng, Guo, Wu, Zhou, and
  Lu]{zhuo2026streaming}
Dong Zhuo, Wenzhao Zheng, Jiahe Guo, Yuqi Wu, Jie Zhou, and Jiwen Lu.
\newblock Streaming visual geometry transformer.
\newblock In \emph{{{ICLR}}}, 2026.

\end{thebibliography}
\end{document}